# MDFI-Net: Multiscale Differential Feature Interaction Network for Accurate Retinal Vessel Segmentation

Yiwang Dong, Xiangyu Deng

*Abstract*—**The accurate segmentation of retinal vessels in fundus images is a great challenge in medical image segmentation tasks due to their highly complex structure distinct from other organs. Currently, deep learning-based methods for retinal vessel segmentation achieved suboptimal outcomes, since vessels with indistinct features are prone to being overlooked in deeper layers of the network. Additionally, the abundance of redundant information in the background poses significant interference to feature extraction, thus increasing the segmentation difficulty. To address this issue, this paper proposes a feature-enhanced interaction network based on DPCN, named MDFI-Net. Specifically, we design a feature enhancement structure, the Deformable-convolutional Pulse Coupling Network (DPCN), to provide an enhanced feature iteration sequence to the segmentation network in a simple and efficient manner. Subsequently, these features will interact within the segmentation network. Moreover, to tackle the problem of feature interaction within the network, we design a simple and lightweight nested skip connection structure inside the M-shaped network and integrate multi-scale subtraction unit (MSU) to facilitate interaction between different hierarchical layers. Finally, considering the grayscale distribution characteristics of a fundus images, we propose a novel loss function calculation method that effectively improves the training effectiveness of retinal vessel areas. Extensive experiments were conducted on publicly available retinal vessel segmentation datasets to validate the effectiveness of our network structure. Experimental results of our algorithm show that the detection accuracy of the retinal blood vessel achieves 97.91%, 97.97% and 98.16% across all datasets. Finally, plentiful experimental results also prove that the proposed MDFI-Net achieves segmentation performance superior to state-of-the-art methods on public datasets.**

## I. INTRODUCTION

THE structure and morphology of retinal vessels in fundus images are closely associated with various human diseases, including glaucoma, diabetic retinopathy, arteriosclerosis, and cirrhosis [1]. Accurate segmentation of retinal blood vessels is a critical step in diagnosing and evaluating the risk of these related diseases. However, manually segmenting retinal vessels requires a high level of skill, which is not only time-consuming and tedious, but may also lead to operator bias, especially when the operator's professional knowledge and ability are insufficient. Therefore, automatic segmentation of retinal vessels is of great value in improving clinical practice and providing diagnostic assistance.

Currently, computer-aided blood vessel segmentation faces significant challenges due to several reasons: 1) Complex biomarkers obstruct the visibility of blood vessels; 2) The vessel has a highly complex morphological structure; 3) The width of blood vessels varies greatly, ranging from 1 pixel to almost 20 pixels [2]; 4) Thin vessels (especially capillaries) exhibit low contrast with the background, and the boundaries are difficult to determine. Fig. 1 shows the complex structures in fundus images and retinal vessels with different scales. To address these challenges, many scholars have devoted considerable efforts. Early research primarily focused on filter design [3-8] and explicit model construction [9-12]. Although these manually crafted extractors can achieve satisfactory results in simple and flat retinal image backgrounds, they may lead to significant segmentation errors in images with complex backgrounds such as structural occlusion. In addition, these feature extractors overly rely on prior knowledge, which means that they have poor robustness. With the widespread application of deep learning in the field of computer vision, many methods based on deep convolutional neural networks (DCNN) have been developed for retinal vessel segmentation [13-15]. In the encoding stage, multi-scale feature extraction [16-21], dilated convolutions [22, 23], and cascaded convolutional layers [24] are often used to alleviate the receptive field limitation caused by fixed convolution kernels. In the decoding stage, many special structures [25] are designed to fully realize feature interaction. Meanwhile, carefully designed attention mechanisms [26-28] are often added to network to alleviate the limitations of convolutional kernels in feature extraction due to the same weight at all pixel positions. In addition, complex skip-connection structures [29-31] are designed to supplement the semantic information missing in the decoder. Although these methods have improved the performance of the network, their ability to reconstruct vascular details is limited, especially for capillaries. In this paper, we propose a Multiscale Differential Feature Interaction Network (MDFI-Net) for retinal vessel segmentation on fundus images. Specifically, we designed a

This work is supported by the National Natural Science Foundation of China (No. 61961037) and the Industrial Support Plan of Education Department of Gansu Province (No. 2021CYZC-30), Xiangyu Deng is the Corresponding author.

Y. Dong and X. Deng are with the College of Physics and Electronic Engineering, Northwest Normal University, 730070 Lanzhou, China (e-mail: dongyw1998@163.com).

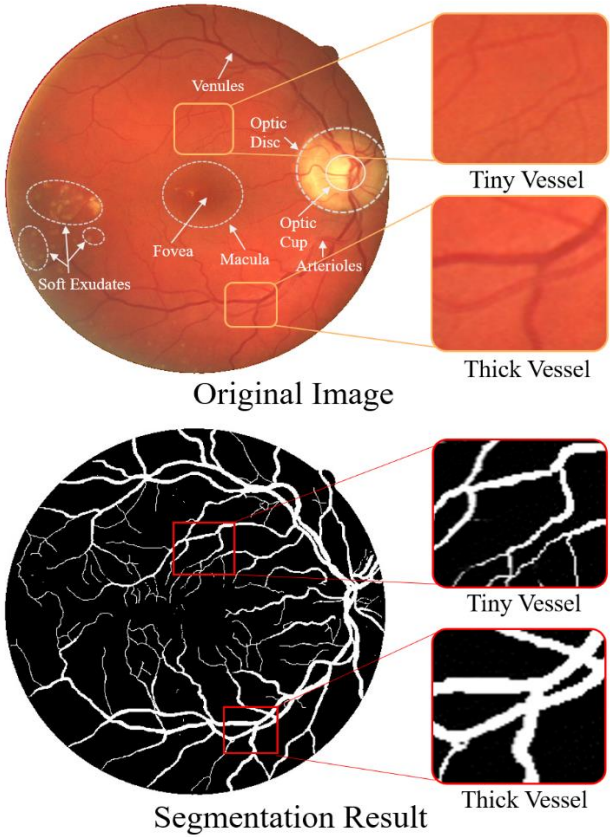

**Fig. 1.** A fundus image with different scale vessels. The non-vascular structures such as optic disc or lesions always affect the segmentation result.

deformable-convolutional pulse coupling network (DPCN) structure and a multi-scale subtraction in M-shaped network (M²Net) to enable the segmentation of different-shaped vessels in low-contrast images. The DPCN structure utilizes a dual-channel coupling method to modulate the influence of surrounding pixels on the central pixel, resulting in varying degrees of enhancement or attenuation for pixels of different gray levels. This approach increases the gray-level contrast between blood vessels and the background, ensuring that the spatial information of small vessels is not easily lost during down-sampling. In M²Net, we integrate the Multi-scale Subtraction Unit (MSU) into a novel skip connection structure to facilitate the interaction of features extracted at multiple scales. The MSU combines multi-scale convolution and element-wise subtraction operations. The redesigned skip connection structure enables the extraction of differential information between different layers and superimpose it onto the initial probabilistic features of each layer. This allows the network to integrate both micro-level texture details and macro-level structural features. To address the issue of class imbalance in fundus images, we have designed a novel loss function to mitigate the problem of missed segmentations. The proposed MDFI-Net model has been evaluated on public datasets — DRIVE, STARE and CHASE_DB1. The results demonstrate that MDFI-Net achieves superior performance in retinal vessel segmentation compared to other models.

The main contributions of this work are as follows: (1) We propose a Deformable-convolutional Pulse Coupling Network (DPCN) structure, and replace traditional convolutions with deformable convolutions to enhance adaptability to the curved structures of blood vessels; (2) We develop a novel skip connection method and integrated the Multi-scale Subtraction Unit (MSU) to facilitate information interaction across different levels; (3) We design an M²Net with deep supervision to segment retinal vessels in various fundus images; (4) We introduce a novel loss function to mitigate the effects of class imbalance in fundus images.

The rest of the paper is organized as follows: Section II briefly reviews related work, Section III introduces the detailed methodology of the proposed MDFI-Net, Section IV shows the experiments, Section V shows the comparison results, Section VI provides discussions about the proposed model, Section VII is the conclusion of our work.

## II. RELATED WORKS

### A. Vessel Segmentation

The goal of retinal vessel segmentation is to accurately delineate vessel boundaries and capture the complete vessel structures. This can mainly be achieved through two broad categories of existing methods: 1) topology-preserving methods focused on maintaining the integrity of vessel structures, and 2) semantic segmentation methods aimed at achieving pixel-level accuracy.

Topology-preserving methods dedicate themselves to maintaining the high integrity of vessel structures. These methods include both implicit [32-34] and explicit [35-37] approaches. For instance, Mosinska et al. [32] employ pre-trained VGG networks to implicitly encode the topological relationships of objects, but its performance is limited. Shit et al. [33] implicitly preserve vessel topology by evaluating the similarity between predicted vessel centerlines and the ground truth centerlines. Hu et al. [35] introduced a topology-preserving loss function to mitigate disconnections and enhance local topological effects. In post-processing, Damseh et al. [37] propose a geometry contraction algorithm to refine the topological structure. However, these methods may sometimes overlook the semantic information of vessels in fundus images, potentially leading to mis-segmentation of occluding structures such as the optic disc. Additionally, some topology-based segmentation algorithms may require substantial computational resources and time, especially for high-resolution retinal images, which could limit their practical application. Lastly, the performance of these methods heavily relies on parameter selection, thereby restricting their robustness and generalization capability.

Semantic segmentation methods primarily utilize an encoder-decoder architecture, focusing on integrating contextual features [16, 18, 38-41] or feature enhancement methods [42-44] to improve the accuracy of pixel-wise classification. For instance, some approaches [16, 18, 22-24, 38, 46] gather global information from larger receptive fields using techniques like dilated convolutions [16, 22, 23, 38] or cascaded convolutions [24], to learn the correlation between pixels in different regions. Multi-scale convolutional kernels [16-21] have been used to extract vessel features of varying thicknesses. Ding et al. [41] employs a High-order Attention module to generate a global attention probability map, modeling relationships between all pixels in the fundus image. [16-18] address the limitations of convolution operations by designing attention mechanisms to weigh the features obtained from convolutions. CS2Net [39] combines local and global features dependencies by introducing spatial and channel attention mechanisms. [29-31] supplements the lost information in the decoder by modifying the skip connection method. Additionally, feature enhancement methods primarily leverage the geometric structural information of vessels. Shi et al. [25] designs affinity feature enhancement networks to ensure sufficient interaction of feature information during the decoding process. Ma et al. [42] enhance vessel features by learning segmentation maps at the level of centerlines to represent complex geometric structures. [43] and [44] enhance vessel edge feature representation by extracting vessel boundaries as prior structures. Inspired by these methods, this study aims to develop a vessel-specific feature enhancement network to selectively enhance vessel features and achieve effective interaction of feature information for precise segmentation within the encoder-decoder network.

### B. Feature Enhancement

By enhancing the features of retinal blood vessels in fundus images, it becomes possible to markedly improve the segmentation of tiny vessel leaks and reduce errors in vessel boundary segmentation, especially in low-contrast images. In the early years, many methods focus on enhancing vessel contrast through filter design. Xia et al. [47] utilize Laplace filters, high-frequency filters, and residual filters to reconstruct regions of interest. On the basis of multi-scale image filtering methods, Min et al. [48] propose a method that can adaptively determine the optimal scale parameter for each pixel during contour evolution to achieve image feature enhancement. However, these methods heavily rely on prior knowledge and parameter settings. Currently, many methods focus on contrast-invariant handcrafted features. Kassim [49] et al. employ a random forest algorithm in conjunction with handcrafted features and U-Net [45] regression prediction results to improve segmentation performance. Recently, Li et al. [50] propose a convex-regional-based gradient model that uses contextually related regional information to locate and enhance vessel features. Additionally, numerous techniques aim to enhance vessel boundaries. Li et al. [43] and Zhang et al. [44] improve vessel edge feature representation by extracting vessel boundaries as prior structures. In VCFNet [51], cross-fusion and parameter dimensionality reduction are used to extract multi-level image edge features. WNet [52] sharpens boundaries by incorporating the prior location and shape information of organs.

In existing vessel features enhance methods, algorithms typically rely either on manual parameter settings or extensive training, which makes them unsuitable for providing enhanced features to segmentation networks. In this paper, we propose a DPCN architecture, aiming to adaptively enhance vessel features across various curved structures and provide more detailed vessel feature.

### C. Feature Interaction

The interaction of feature information within neural networks not only helps in capturing complex correlations within data but also generates more robust feature representations. Therefore, the interaction between different features within the network is crucial for segmentation performance before the feature information is decoded into target results. Currently, some common techniques to promote feature interaction include multi-channel coupling [17, 21, 25, 26], multi-path fusion [18, 28], dense connection structures [53, 54], and improved skip connections [29-31, 55-59]. Several methods have been proposed to fuse feature information across different scales. For instance, FA-Net [26] assigns weights to vessel features at different scales using attention mechanisms, while CSC-Net [17] utilizes an adaptive feature fusion module (AFF) to merge multi-scale vessel features. Additionally, CSU-net [18] separately designs the context channel block and spatial channel block branches, and utilizes a feature fusion module (FFM) to combine the features extracted from two paths. Inspired by DenseNet [60], MSD-Net [53], and MSI-MFNet [54] combine multiscale feature maps and dense connections to produce high-level feature representations. Compared to skip connection structures, most of these methods rely on attention mechanisms to promote interaction between features within the same layer, overlooking the semantic gap between the encoder and decoder. In medical image segmentation tasks, Unet++ [29] initially emphasizes the importance of skip connection structures for fine segmentation and designs a nested skip connection mechanism. Subsequently, various variants of skip connections continue to be proposed, such as Unet3+ [30] and Unet# [55], which achieve different interaction effects by changing the number and density of skip connections. Attention Unet [31] and Attention Unet++ [56] achieve encoder-decoder feature interaction by introducing attention mechanisms into skip connections. Compared to mere intra-layer interaction, inter-layer information interaction is more instructive. DCA-Unet [59] integrates feature information between all layers through the Dual Cross-Attention block and then distributes the results to each layer of the decoder. Inspired by the above methods, we propose a segmentation network called M²Net. The network adopts an M-shaped architecture and incorporates redesigned skip connection structures to promote feature interaction between different layers. Meanwhile, we integrate MSU units into skip connections to eliminate redundant information between different layers and propagate differential information to the decoder.

### III. METHOD

#### A. Network Architecture

As shown in Fig. 2, this paper introduces a segmentation model for retinal vessels, which mainly consists of two parts: a feature extraction iteration module named Deformable-convolutional Pulse Coupling Network (DPCN) and a M-shaped segmentation network with Multi-scale Subtraction Unit (M²Net). In the proposed network, it introduces deformable convolutional network (DCN) and multi-scale subtraction unit (MSU) to further improve the result performance. Within the feature extraction iteration module, we incorporate a deformable convolutional network, which helps in shaping the receptive field, thereby greatly enhancing the network's flexibility and adaptability to various vessel structures. The segmentation network is designed with encoder-decoder structure, in which we adapt Res2Net-50 as the backbone of encoder to capture five hierarchical features, and the decoder consists of a four-layer structure to reconstruct pixel-wise semantic features. At the outer sides of the encoder and decoder, we add additional down-sampling and up-sampling paths, to help reduce pixel level information loss in the network. Additionally, we incorporate nested skip connections between each encoding and its corresponding decoding layers. This approach not only alleviates the spatial loss of encoders caused by pooling operations but also reduces the semantic gap between the feature maps of the encoder and decoder sub-networks. As the intricate skip connections may increase the risk of overfitting, we use sparse linking methods and embed a multi-scale subtraction unit (MSU) in each layer, which can effectively capture vascular information at different scales while improving the recognition ability of vascular boundaries. Finally, a crafted joint loss function is employed to address the proportional difference between foreground and background, to enhancing the network's generalization capability.

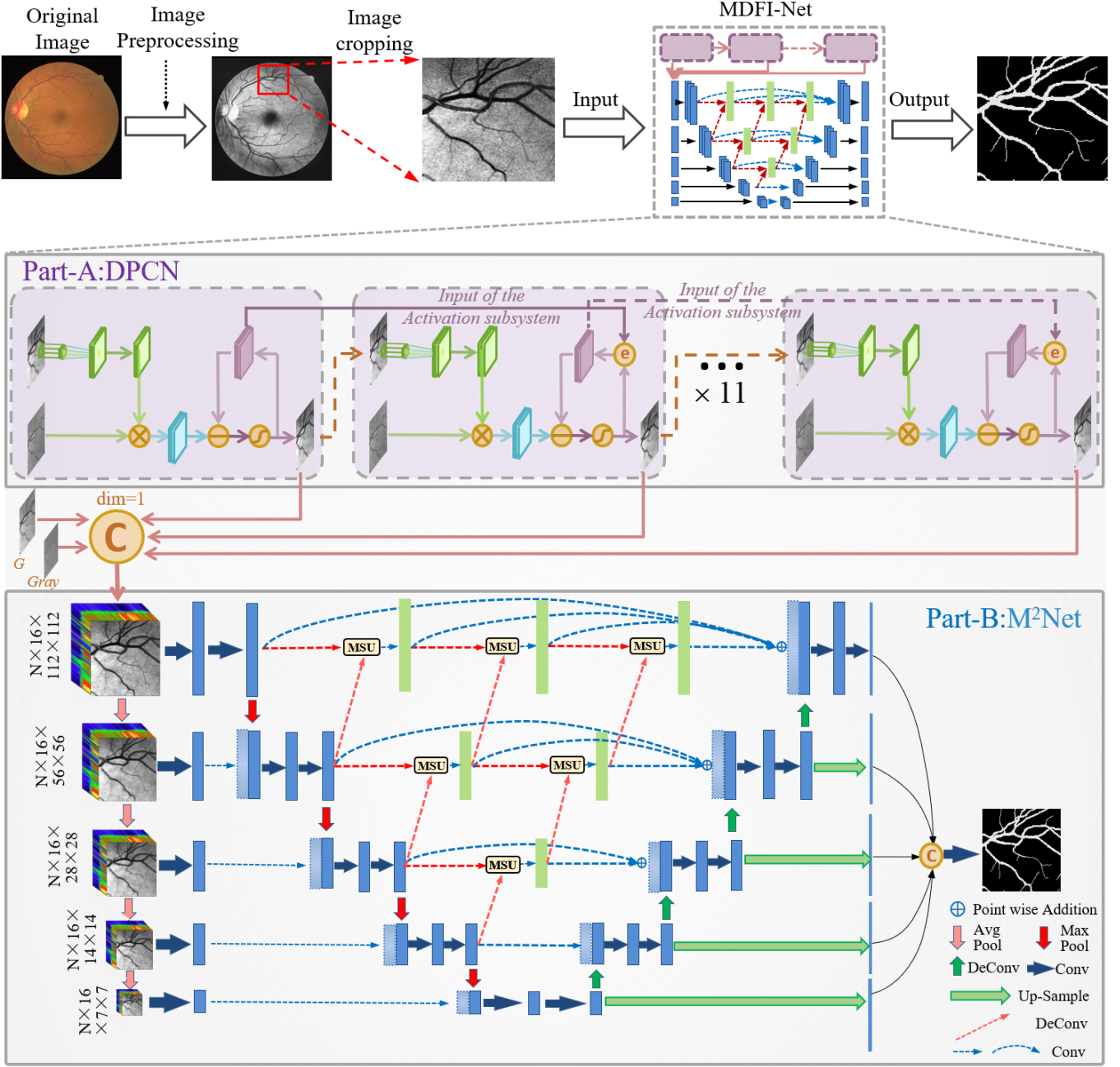

Fig. 2. The architecture of the proposed MDFI-Net consists of two components: part-A, DPCN, and part-B, M²Net. In part-A, a Deformable-convolutional Pulse Coupling Network (DPCN) is designed for feature enhancement, which utilizes the coupling effect between surrounding pixels and central pixels to iteratively enhance target blood vessels with low contrast, in order to obtain pixel level vascular feature information. In part-B, an M²Net is designed for precise segmentation of retinal vessels. We incorporate a nested skip connection structure inside it to facilitate feature interaction between different layers and reduce semantic loss during the decoding process.

## B. Deformable-convolutional Pulse Coupling Network

In convolutional networks, deep semantic features are typically derived from shallow features through convolutional operations, making the effective extraction of shallow features crucial for the accuracy of segmentation results. In retinal images, blood vessels exhibit similarities with surrounding tissues, and the image contrast is low. To effectively extract vessel information at the network's shallow layers, we propose Deformable-convolutional Pulse Coupling Network (DPCN), which consists of a series of iterative submodules as shown in Fig. 2 Part-A.

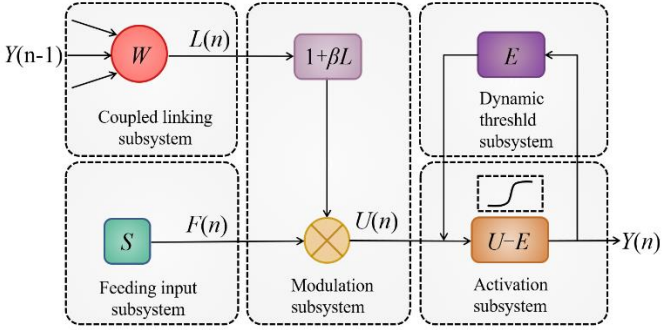

**Fig. 3.** A detailed depiction of one iteration process in DPCN. The iteration module consists of five subsystems. The Coupled linking subsystem and Modulation subsystem work together to provide enhancement for the input pixels, while the Dynamic threshold subsystem and Activation subsystem jointly adjust the threshold value to achieve appropriate attenuation for different pixels

During one iteration of the DPCN, the input information undergoes processing through five subsystems, including the Coupled Linking Subsystem, the Feeding Input Subsystem, the Modulation Subsystem, the Dynamic Threshold Subsystem, and the Activation Subsystem. The Coupled Linking subsystem and Modulation subsystem work together to provide enhancement for the input pixels, while the Dynamic Threshold subsystem and Activation subsystem jointly adjust the threshold value to achieve appropriate attenuation for different pixels. This method allows different grayscale features to be enhanced to varying degrees during the continuous iterative process, while also providing modulation guidance for the next iteration. The entire iterative process is expressed as follows:

$$Y(n) = Sigmoid\left(F(n) \cdot \left(1 + \beta \cdot L(n)\right) - E(n)\right) \quad (1)$$

here, Y(n) represents the result of the $n^{th}$ iteration of the entire system. F(n), L(n), and E(n) respectively denote the results of the Feeding Input Subsystem, Coupled Linking Subsystem, and Dynamic Threshold Subsystem during the $n^{th}$ iteration.

To elucidate the operational mechanism of the proposed module, we provide a detailed depiction of one iteration process in Fig. 3. This model comprises five subsystems, each of which is expressed as follows:

$$L(n) = Conv\left(Y(n-1)\right) \quad (2)$$

$$F(n) = I_{OR} \quad (3)$$

$$U(n) = F(n) \cdot \left(1 + \beta \cdot L(n)\right) \quad (4)$$

$$E(n) = e^{-a_E} E(n-1) + V_E Y(n-1) \quad (5)$$

$$Y(n) = Sigmoid\left(U(n) - E(n)\right) \quad (6)$$

Eqs. (2) represents the Coupled Linking Subsystem, which simulates the influence of inputs from other neurons on the current neuron. This is manifested as the coupling effect of surrounding pixels on the center pixel, where $Conv(\cdot)$ denotes the deformable convolution layer, which we will introduce in detail later. Eqs. (3) represents the Feeding Input Subsystem, which simulates the input of the current neuron, namely the grayscale values of the pixels, where $I_{OR}$ denotes the original image. Eqs. (4) represents the Modulation Subsystem, in which the states of the feeding units and linking units combine in a second-order manner to produce the internal state $U(n)$ of the neuron, with the degree of combination controlled by the coefficient $\beta$. Eqs. (5) represents the Dynamic Threshold Subsystem, which simulates the variation process of the internal threshold of the neuron. Throughout the iterative process, the internal threshold of the neuron dynamically changes, with its growth and decay regulated by hyperparameters $a_E$ and $V_E$. When the neuron's activity in the previous iteration cycle is high (i.e., when the result of $Y(n-1)$ is relatively large), the threshold will experience a significant increase and then begin to gradually decay. Eqs. (6) represents the Activation Subsystem, which simulates the activation process of the neuron. Here, we do not use a step function for simulation because we aim to obtain enhanced feature maps with prominent grayscale differences.

To accommodate the characteristic trends of blood vessels, we have introduced a deformable convolutional network in the modulation subsystem, which modifies the traditional convolution operation by incorporating learnable offsets as shown in Fig. 4. Compared with ordinary convolution, these learnable offsets enable the convolutional kernel to adjust its receptive field, aligning it more closely with the actual shape of the target object, and capable to select capturing features within the region of interest on the feature map. In the deformable convolution network, convolutional layers are trained to learn offsets from input feature maps, which are used to compute corresponding displacements. These displacements, when added to the pixel indices of the original image, yield adjusted indices. To ensure smooth backpropagation, the pixel values corresponding to these adjusted indices are obtained via bilinear interpolation. Finally, this process results in convolving the displaced pixels to generate the output. The computational formula of the deformable convolution is:

$$F(m,n) = \sum_{i=-1}^{1} \sum_{j=-1}^{1} W(i,j) \cdot x\left(m + (i + \Delta m), n + (j + \Delta n)\right) \quad (7)$$

where $i, j$ represents the position of the corresponding convolution kernel, and $i, j \in \{-1, 0, 1\}$, $\Delta m, \Delta n$ represents the result of offset result.

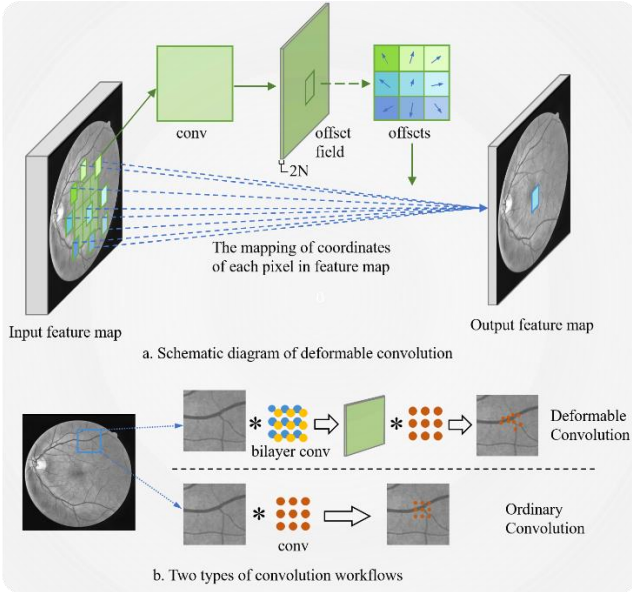

**Fig. 4.** Workflow of deformable convolution. The introduction of deformable convolution can help enhance the adaptability of network to the curved structures of retinal vessels.

### C. M-shaped Segmentation Network with Multi-scale Subtraction Unit

As shown in Fig. 2 Part-B, we propose M²Net for vascular segmentation, which introduces a spatial dimensionality reduction mechanism based on the original U-shaped architecture and improves the skip connection method by introducing Multi-scale Subtraction Units (MSU) on the basis of nested skip connections. Within the MSU, multi-scale convolution operations are dedicated to extracting information of different scales from feature maps, which helps alleviate the problem of feature weight singularity caused by a single scale convolution kernel, thus enabling attention to targets of different scales. In addition, the subtraction operation can extract differential features from different branches, and in the nested skip connection structure we designed, all differential information will be aggregated onto the final feature map of each layer to prevent the loss of detailed information.

The Multi-scale Subtraction Unit (MSU) plays a crucial role as a feature fusion and extraction module, as illustrated in Fig. 5. MSU utilizes convolution filters of different scales with size 1×1, 3×3, and 5×5, to extract multi-scale feature information, and integrates subtraction operations to compute detail and structural disparity values. In convolutional networks, the convolution kernels act as receptive fields, enabling deeper perception and information capture of either the original image or higher-level image feature information. The larger convolution kernels can capture more global information, such as higher semantic feature levels, whereas the smaller kernels capture finer and more detailed information. Considering the variation thickness in retinal vessels, employing multi-scale convolutions not only accommodates structural differences in retinal vessels but also increases richness of feature representation. Moreover, by establishing differential relationships between input information, the subtraction operation within MSU can effectively suppress redundancy, offering notable efficiency compared to the additive operations in traditional networks, which is particularly advantageous given the presence of a significant proportion of irrelevant background information in retinal images. The entire MSU process can be formulated as:

$$\begin{aligned}
F_{out} = Conv( \ &| Conv(F_A)_{1\times1} - Conv(F_B)_{1\times1} | + \\
&| Conv(F_A)_{3\times3} - Conv(F_B)_{3\times3} | + \\
&| Conv(F_A)_{5\times5} - Conv(F_B)_{5\times5} | \ )
\end{aligned} \quad (8)$$

where $Conv(\cdot)_{n\times n}$ represents the convolutional filter of size $n \times n$. The MSU extracts features of $F_A$ and $F_B$ at different scales and effectively focuses on their differences in texture and structure through its adept subtraction operations, thereby providing the decoder with richer information.

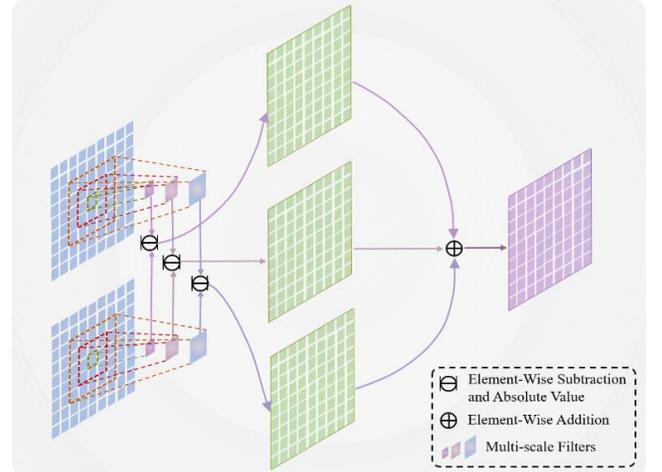

**Fig. 5.** MSU employs a subtraction operation to extract differential features from two branches, which are then overlaid onto the initial feature maps at each layer. Meanwhile, the internal multi-scale convolution operation helps alleviate the problem of feature weight singularity caused by a single scale convolution kernel.

Furthermore, a well-designed skip connection structure is crucial for the segmentation results, especially in segmentation networks with an encoder-decoder architecture. Skip connections have demonstrated their effectiveness in recovering fine-grained details of target objects, enabling the generation of segmentation masks with intricate details even against complex backgrounds. However, an excessively dense skip connection structure not only increases the network's complexity and computational burden but also renders the network challenging to train, leading to potential overfitting issues, particularly detrimental in scenarios with limited medical image data. Therefore, we propose a novel skip pathway built upon nested connections, aiming to alleviate information loss and enhance segmentation accuracy.

As illustrated in Fig.2 Part-B, MSUs are horizontally and vertically arranged within the redesigned skip pathways to capture a series of differential features from various levels. As indicated by the red path, MSUs leverage their internal multi-scale convolution filters and subtraction operations to extract distinctive features from different hierarchical levels. These differential features (shown in green stripe) are then aggregated onto each layer's original feature map (shown in blue stripe) through element-wise addition. In this way, the redesigned skip connections not only tighten the interconnection between different levels of information but also optimize gradient propagation. Moreover, the differential information between different levels is accumulated onto the initial feature maps of each layer to generate complementarity-enhanced features, which is highly beneficial, especially in scenarios of insufficient image feature contrast.

### D. Loss Function

MDFI-Net is an end-to-end deep learning network. The purpose of training the network is to predict whether each pixel is a vessel, which essentially transforms the segmentation task into a pixel-level classification task.

However, there is a significant imbalance between the number of vessel pixels and background pixels in fundus images. According to our statistics, the number of background pixels is approximately 10.6 to 13.3 times that of vessel pixels across different datasets. To address this imbalance, we designed a combined loss function with adjusted weights to optimize the network parameters during the training stage, thereby increasing the network's focus on vessel regions.

The mathematical formula is shown in (9), as shown at the bottom of this page. In this formula, $W_k$ represents the adjusted weight, while $k$ denotes the class, either background or vessel. $N$ represents the total number of pixels of a certain class in the image, and $i,j$ represent the pixel positions. When $k = 1$, the formula calculates the loss for all vessel pixels in the image based on their positional information in the labels. Conversely, when $k = 0$, it calculates the loss for the background pixels. Clearly, this requires all input values to be between 0 and 1. Therefore, to ensure the network's output falls within this range, we typically add a sigmoid function at the output layer.

## IV. Experiment

### A. Datasets

Three public fundus image datasets are used in our experiment, which were captured by different devices and with different image sizes, the overview information is given in TABLE I.

### TABLE I
### An Overview of Fundus Images Datasets Used for This Study

| Dataset | Quantity | Training-Testing Split | Resolution |
|---------|----------|------------------------|------------|
| DRIVE | 40 | 20-20 | 565×584 |
| STARE | 20 | Leave-one-Out | 700×605 |
| CHASE_DB1 | 28 | 20-8 | 999×960 |

DRIVE: The Digital Retinal Images for Vessel Extraction (DRIVE) dataset[1] contains 40 fundus images, which have been equally divided into training set and testing set, and a circular field of view (Fov) mask with a diameter of approximately 540 pixels accompanies each image within both sets. Each image has a resolution of 565×584 pixels with eight bits per color channel (3 channels). In the testing set, there are two experts' manual segmentations, the segmentation produced by the first expert is used as the ground truth, and the second expert's segmentation will be used as manual results and compared with various algorithms.

STARE: The Structured Analysis of the Retina (STARE) dataset[2] contains 20 color fundus images with the same resolution as 700×605 pixels. Each image in the dataset has two corresponding manual segmentation maps used to label retinal vessels. In order to conduct a fair performance comparison, we use the leave-one-out method for cross-validation as used in most literatures.

CHASE_DB1: CHASE_DB1[3] is a retinal images subset of the Child Heart and Health Study in England (CHASE) with 28 retinal images captured from both eyes of 14 children. Each image has a resolution of 999×960 pixels and two manual segmentations by experienced ophthalmologists. In our experiment, we divided the dataset into 20 cases for training and 8 cases for testing.

### B. Implementation details

Experimental setup: All the models were implemented in pytorch 1.13.1. Experiments were performed on win10 (64-bit) operating system, the processor of the running platform is Intel® Core™ i7-10510U CPU @ 4.20 GHz, and GPU is NVIDIA RTX A4000 with 16G memory.

$$\ell_W = \sum_{k=0}^{1} W_k \left[ \frac{1}{2} \cdot \left( 1 - \left( \frac{2 \cdot \sum\limits_{i,j}^{N} \left( y_{i,j}^k - \hat{y_{i,j}^k} \right)}{\sum\limits_{i,j}^{N} y_{i,j}^k + \sum\limits_{i,j}^{N} \hat{y_{i,j}^k}} \right) \right) + \frac{1}{2} \cdot \frac{1}{N} \sum_{i,j}^{N} \left( g_{i,j}^k \log p_{i,j}^k + \left( 1 - g_{i,j}^k \right) \log \left( 1 - p_{i,j}^k \right) \right) \right] \tag{9}$$

*Parameter settings:* In our network, we leveraged the max-pooling layer for down-sampling and the bilinear interpolation for up-sampling. In the output layer, we employed a 1×1 convolutional layer to change the number of channels and used the Sigmoid function to generate the probability output. Except for that, all activation functions were set to ReLU. In addition, we adopted the Adam algorithm as the optimizer with a learning rate of $lr$ = 1e-4 and set the batch size = 8. Considering the proportion of positive and negative samples in fundus images, we set the parameters $W_0$ = 0.9 and $W_1$ = 0.1 in the loss function.

*Pre-processing:* In our experiment, each fundus image was preprocessed in three steps (see Fig. 6). Firstly, since the information of blood vessel contained in green channel has low noise and high contrast, which is much better than the red and blue channels, we mixed the R/G/B channels into a grayscale image at an optimal ratio of 0.2793:0.7041:0.0166 [61]. Secondly, to improve the contrast of all pixels evenly, we employed the *contrast* limited adaptive histogram equalization (CLAHE) method to process each mixed grayscale image. Finally, we applied the gamma correction to adjust the ratio of dark part and bright part in the image, which makes the vessel more clear.

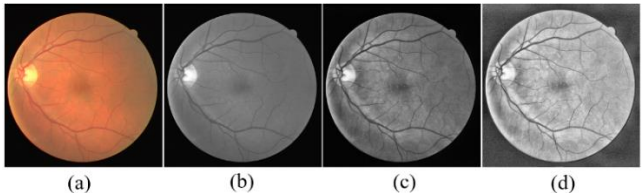

(a)　　　　　(b)　　　　　(c)　　　　　(d)

**Fig. 6.** Preprocessing steps on datasets. (a) shows the original images, (b) is the result of mixing RGB channels, (c) shows the result of CLAHE based on the previous operation, (d) is gamma correction followed by CLAHE.

## C. Evaluation Metrics

Retinal vessel segmentation aims to categorize each pixel in a fundus image as either a vessel or non-vessel, which is a binary classification task. The proposed model of this work produces a probability map, which illustrates the likelihood of each pixel being classified into the vessel category. In the experiment, we employed a threshold of 0.5 to constrain the output probabilities and convert them into binary results. In the vessel category, pixels are labeled true positives (TP) if correctly classified, otherwise, they are labeled as false positives (FP). Conversely, in the non-vessel category, pixels are labeled true negatives (TN) if accurately classified; otherwise, they are labeled as false negatives (FN). In order to conduct a fair performance comparison, we introduced five evaluation metrics, including ACC, SEN, SPE, F1, and AUC, which are listed in TABLE II. Among these, the accuracy (ACC) represents the proportion of correctly classified pixels across all categories, commonly used to evaluate the segmentation effectiveness of a model. The sensitivity (SEN) denotes the proportion of correctly classified pixels in the vessel category, indicating the model's ability to identify positive samples. The specificity (SPE) represents the proportion of correctly classified pixels in the background category, illustrating the model's ability to recognize negative samples. Additionally, the F1_score (F1) combines SEN and SPE, offering a comprehensive assessment of the model's performance in binary classification tasks. Correspondingly, the AUC stands for the Area Under the ROC Curve, with a larger value indicating better overall performance of the model. Unless otherwise specified, performance metrics were calculated using only pixels within the field of view (FOV).

TABLE II
METRICS FOR EVALUATION IN OUR WORK

| Metrics | Description |
|---|---|
| ACC (accuracy) | ACC = (TP + TN) / (TP + TN + FP + FN) |
| SEN (sensitivity) | SEN = TP / (TP + FN) |
| SPE (specificity) | SPE = TN / (TN + FP) |
| F1 (F1_score) | F1 = (2×TP) / (2×TP + FP + FN) |
| AUC | Area Under the ROC Curve |

## V. RESULT

### A. The Segmentation Results of The Proposed Model

This paper employs five performance evaluation metrics: ACC (accuracy), SEN (sensitivity), SPE (specificity), F1 Score (F1), and Area Under the Receiver Operating Characteristic Curve (AUC) to objectively quantify the proposed model. To contrast with manually segmented results, we report the results of MDIF-Net on *three* public datasets (DRIVE, STARE and CHASE_DB1), the experimental results are summarized in TABLE III. Based on detailed comparison in TABLE III, the performance evaluation indexes of our algorithm are better than the manually segmented results.

Furthermore, we illustrate the segmentation result of the proposed MDIF-Net on fundus image datasets (as shown in Fig. 7). The heatmap result indicates the likelihood of retinal vessels, where higher values represent higher probabilities. In the original image, the diameter of retinal vessels varies a lot, and the intensity contrast is extremely weak, especially for small vessels. However, from Fig. 7, we can see most vessels are effectively segmented by MDIF-Net, and even tiny vessels are assigned high probabilities in the heatmap.

TABLE III
PERFORMANCE MEASURES FOR THE **DRIVE**, **STARE** AND **CHASE_DB1** DATASETS.

| Dataset | Method | ACC | SEN | SPE | F1 | AUC |
|---|---|---|---|---|---|---|
| DRIVE | 2nd observer | 0.9473 | 0.7760 | 0.9725 | 0.7881 | – |
| | Ours (Fov)* | 0.9652 | 0.8369 | 0.9779 | 0.8379 | 0.9849 |
| | Ours (All) | 0.9791 | 0.8366 | 0.9883 | 0.8379 | 0.9897 |
| STARE | 2nd observer | 0.9346 | 0.8956 | 0.9381 | 0.7417 | – |
| | Ours (Fov)* | 0.9705 | 0.8661 | 0.9864 | 0.8638 | 0.9913 |
| | Ours (All) | 0.9797 | 0.8528 | 0.9923 | 0.8581 | 0.9928 |
| CHASE_DB1 | 2nd observer | 0.9545 | 0.8105 | 0.9711 | 0.7963 | – |
| | Ours (Fov)* | 0.9687 | 0.9031 | 0.9802 | 0.8358 | 0.9891 |
| | Ours (All) | 0.9816 | 0.9029 | 0.9898 | 0.8358 | 0.9924 |

*Field of view (Fov) is defined for performance measurement. Fov denotes that only the internal pixels of the fundus are used for performance evaluation, while All indicates that all pixels in the images are taken into consideration.

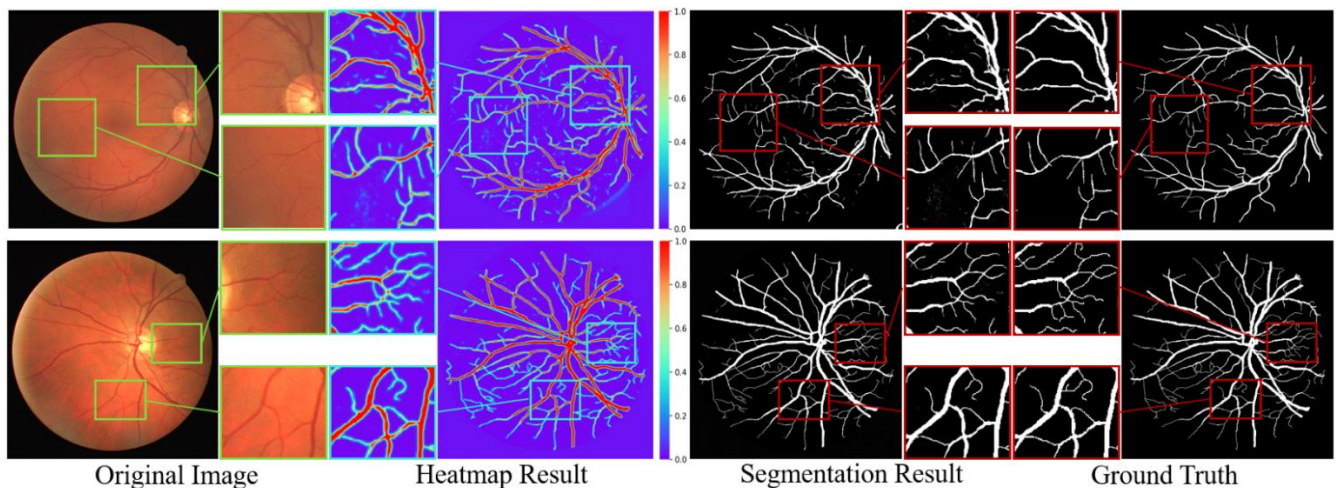

**Fig. 7.** Examples of segmentation results in fundus images. From left to right is the original image with various zoomed-in patches of different scale retinal vessels; the heatmap result of the proposed MDIF-Net; the final segmentation result with threshold of 0.5, and the ground truth.

*B. Ablation Analysis*

To demonstrate the effectiveness of different network architectures and modules, a comparative ablation experiment was conducted on the DRIVE dataset (as shown in Table IV). We validated the contribution of the nested skip connections embed with MSU by comparing the results of M-Net and M²Net. The novel skip connections make a significant improvement in SEN (sensitivity), which indicates the proportion of correctly predicted pixels in the vessel area, demonstrating the MSU's capability to capture subtle differential features. After adding the DPCN module, ACC (accuracy) and SPE (specificity) rate both improved, which may be due to the redundant information of background being eliminated by the DPCN module. Through this mechanism, the DPCN module helps to concentrate network's attention on the target vessels while reducing interference from background, so the performance is further improved.

TABLE IV
ABLATION STUDY OF THE PROPOSED METHOD ON **DRIVE** DATASET

| M-Net | DPCN | NSC* | ACC | SEN | SPE | F1 | AUC |
|---|---|---|---|---|---|---|---|
| √ | | | 0.9588 | 0.8091 | 0.9773 | 0.8227 | 0.9753 |
| √ | √ | | 0.9704 | 0.8137 | 0.9864 | 0.8254 | 0.9824 |
| √ | | √ | 0.9663 | 0.8258 | 0.9808 | 0.8250 | 0.9838 |
| √ | √ | √ | 0.9791 | 0.8366 | 0.9883 | 0.8379 | 0.9897 |

*NSC stands for the nested skip connections embedded with MSU

*C. Comparing to existing methods*

We conduct a comparative analysis between our proposed method and several contemporary state-of-the-art methodologies. Evaluations are performed across three publicly available datasets, and the results are summarized in TABLE V, VI. The performance metrics of existing methods are obtained from the original papers. Furthermore, the methods marked with stars utilize Fov for metric computation, while those without stars are not explicitly specify their employment of Fov. Here, our study reports the performance of the proposed method under both Fov and without Fov conditions. For fairness, our method used the same data split as the other methods.

Among the introduced evaluation metrics, AUC is a significant performance indicator, especially when applied to image segmentation tasks. Unlike other metrics, AUC can assess the segmentation quality of probability maps by using different thresholds, and provides a more comprehensive evaluation result compared to single-threshold metrics. Besides, the F1 score is often used to analyze segmentation performance, particularly in cases of class imbalance, which is especially relevant in retinal vessel segmentation tasks, where the number of background pixels in fundus images greatly exceed the vessel pixels. For all datasets, the proposed method demonstrates superior performance in two comparative metrics (AUC and F1), which indicates that the constructed network is optimal in overall performance. Additionally, the proposed method also outperforms other methods in SEN (sensitivity), which suggests that the network is more effective in detecting vessels. In fundus images, non-vessel regions are significantly larger than vessel regions. Therefore, ACC (accuracy) is more persuasive than SPE (specificity), as it strikes a balance between SEN (sensitivity) and SPE (specificity). The ACC of proposed method is also superior than other methods. From TABLE V, it can be observed that the proposed method performs slightly lower on the SPE index compared to the AA-WGAN method. We believe that this may be due to the AA-WGAN method being more sensitive to TN, thus increasing its performance on the SPE index. It also can be observed that, the manual segmentation result of the second observer demonstrates significant performance on the SEN index. This is due to the different segmentation standards of blood vessels, which results in over segmentation on the fundus image. That is, a large number of blood vessels have been accurately segmented, but correspondingly, many background pixels have been mistakenly classified as blood vessels.

TABLE V
COMPARISON WITH EXISTING METHOD ON THE **DRIVE** AND **STARE** DATASETS

| Method type | Method | Year | DRIVE | | | | | STARE | | | | |
|---|---|---|---|---|---|---|---|---|---|---|---|---|
| | | | ACC | SEN | SPE | F1 | AUC | ACC | SEN | SPE | F1 | AUC |
| Manual | 2nd observer * | – | 0.9473 | 0.7760 | 0.9725 | 0.7881 | – | 0.9346 | **0.8956** | 0.9381 | 0.7417 | – |
| Unsupervised methods | Upadhyay et al. [62] * | 2020 | 0.9560 | 0.7890 | 0.9720 | 0.7.60 | 0.9610 | 0.9610 | 0.7360 | 0.9810 | – | 0.9660 |
| | Palanivel et al. [63] | 2021 | 0.9480 | 0.7375 | 0.9788 | – | 0.9590 | 0.9542 | 0.7484 | 0.9780 | – | 0.9711 |
| | Khan et al. [65] | 2021 | 0.9610 | 0.8125 | 0.9763 | – | 0.9730 | 0.9586 | 0.8078 | 0.9721 | – | 0.9782 |
| Supervised methods | U-Net [45] | 2015 | 0.9863 | 0.7769 | 0.9866 | 0.8109 | 0.9767 | 0.9570 | 0.6241 | **0.9946** | 0.7469 | 0.9612 |
| | R2U-Net [64] | 2018 | 0.9556 | 0.7792 | 0.9813 | 0.8171 | 0.9784 | 0.9634 | 0.7756 | 0.9820 | 0.7928 | 0.9815 |
| | IterNet [66] * | 2019 | 0.9573 | 0.7735 | 0.9838 | 0.8205 | 0.9816 | 0.9701 | 0.7715 | 0.9886 | 0.8146 | 0.9881 |
| | FA-Net [26] | 2020 | <u>0.9769</u> | 0.8145 | <u>0.9883</u> | – | <u>0.9895</u> | **0.9797** | 0.8505 | 0.9889 | – | <u>0.9924</u> |
| | HA-Net [27] | 2020 | 0.9581 | 0.7991 | 0.9813 | 0.8293 | 0.9823 | <u>0.9772</u> | 0.8186 | 0.9844 | 0.8379 | 0.9881 |
| | Sine-Net [77] | 2020 | 0.9689 | 0.7987 | 0.9854 | – | 0.9851 | 0.9711 | 0.6776 | **0.9946** | – | 0.9807 |
| | DSCA-Net [45] | 2020 | 0.9571 | 0.8339 | 0.9750 | <u>0.8319</u> | 0.9821 | 0.9664 | 0.8463 | 0.9802 | 0.8384 | 0.9881 |
| | SA-UNet [67] | 2021 | 0.9698 | 0.8212 | 0.9840 | 0.8263 | 0.9864 | 0.9607 | 0.6853 | 0.9607 | 0.7796 | 0.9748 |
| | MPS-Net [70] | 2021 | 0.9563 | <u>0.8361</u> | 0.9740 | 0.8287 | 0.9809 | 0.9689 | 0.8566 | 0.9819 | 0.8491 | 0.9873 |
| | Genetic U-Net [71] * | 2021 | 0.9577 | 0.8300 | 0.9758 | 0.8314 | <u>0.9895</u> | 0.9719 | <u>0.8658</u> | 0.9846 | <u>0.8630</u> | 0.9921 |
| | MFI-Net [72] | 2022 | 0.9699 | 0.8166 | 0.9847 | 0.8249 | 0.9884 | 0.9687 | 0.8220 | 0.9854 | 0.8396 | 0.9897 |
| | CSG-Net [73] | 2022 | 0.9576 | 0.7943 | 0.9814 | 0.8310 | 0.9823 | 0.9692 | 0.8298 | 0.9855 | 0.8493 | 0.9895 |
| | Wave-Net [69] | 2023 | 0.9561 | 0.8164 | 0.9764 | 0.8254 | – | 0.9664 | 0.8284 | 0.9821 | 0.8349 | – |
| | Swin-ASNet [78] | 2023 | 0.9694 | 0.8283 | 0.9829 | – | 0.9863 | 0.9733 | 0.8456 | 0.9821 | – | 0.9883 |
| | AA-WGAN [76] | 2023 | 0.9651 | 0.7923 | **0.9903** | – | 0.9788 | 0.9719 | 0.8234 | 0.9918 | – | 0.9873 |
| | DA-Res2UNet [75] | 2023 | 0.9701 | 0.8200 | 0.9847 | 0.8269 | 0.9910 | 0.9770 | 0.8080 | 0.9885 | 0.8157 | 0.9910 |
| | HFRF-Net [81] | 2024 | 0.9706 | 0.8262 | 0.9826 | 0.8308 | 0.9886 | 0.9656 | 0.7502 | 0.9900 | 0.8160 | 0.9831 |
| | IMFF-Net [82] | 2024 | 0.9621 | 0.8275 | 0.9860 | 0.7977 | – | 0.9707 | 0.8634 | 0.9869 | 0.8347 | – |
| | Ours (Fov) * | 2024 | 0.9652 | **0.8369** | 0.9779 | **0.8379** | 0.9849 | 0.9705 | <u>0.8661</u> | 0.9864 | **0.8638** | 0.9913 |
| | Ours (All) | 2024 | 0.9791 | 0.8366 | 0.9883 | 0.8379 | **0.9897** | **0.9797** | 0.8528 | <u>0.9923</u> | 0.8581 | **0.9928** |

*Field of view (Fov) is defined for performance measurement. Fov denotes that only the internal pixels of the fundus are used for performance evaluation, while All indicates that all pixels in the images are taken into consideration.

TABLE VI
COMPARISON WITH EXISTING METHOD ON THE **CHASE_DB1** DATASET

| Method type | Method | Year | ACC | SEN | SPE | F1 | AUC |
|---|---|---|---|---|---|---|---|
| Manual | 2nd observer * | – | 0.9545 | 0.8105 | 0.9711 | 0.7963 | – |
| Unsupervised methods | Upadhyay et al. [62] * | 2020 | 0.9580 | 0.7540 | 0.9750 | – | 0.9530 |
| | Palanivel et al. [63] | 2021 | 0.9459 | 0.7237 | 0.9703 | – | 0.9592 |
| | Khan et al. [65] | 2021 | 0.9578 | 0.8012 | 0.9730 | – | 0.9708 |
| Supervised methods | U-Net [45] | 2015 | 0.9637 | 0.8138 | 0.9737 | 0.8066 | 0.9793 |
| | IterNet [66] * | 2019 | 0.9655 | 0.7970 | 0.9823 | 0.8073 | 0.9851 |
| | DC-UNet [68] | 2020 | 0.9488 | <u>0.8967</u> | 0.9540 | – | 0.9785 |
| | FA-Net [26] | 2020 | <u>0.9803</u> | 0.8334 | 0.9862 | – | 0.9912 |
| | HANet [27] | 2020 | 0.9670 | 0.8239 | 0.9813 | 0.8191 | 0.9871 |
| | SA-UNet [67] | 2021 | 0.9755 | 0.8573 | 0.9835 | 0.8153 | 0.9905 |
| | Genetic UNet [71] * | 2021 | 0.9667 | 0.8463 | 0.9845 | 0.8223 | 0.9880 |
| | TiM-Net [79] | 2022 | 0.9774 | 0.8261 | 0.9728 | 0.8295 | 0.9857 |
| | MFI Net [72] | 2022 | 0.9770 | 0.8335 | 0.9868 | 0.8205 | <u>0.9916</u> |
| | CSGNet [73] * | 2022 | 0.9681 | 0.7947 | 0.9855 | 0.8247 | 0.9886 |
| | MRC-Net [74] | 2023 | 0.9779 | 0.8485 | 0.9887 | 0.8091 | 0.9857 |
| | MGA-Net [80] | 2023 | 0.9637 | 0.8560 | 0.9771 | 0.8248 | 0.9723 |
| | DA-Res2UNet [75] | 2023 | 0.9761 | 0.8139 | 0.9892 | <u>0.8352</u> | 0.9879 |
| | AA-WGAN [76] | 2023 | 0.9694 | 0.8335 | **0.9914** | – | 0.9781 |
| | HFRF-Net [81] | 2024 | 0.9656 | 0.7502 | 0.9900 | 0.8160 | 0.9831 |
| | IMFF-Net [82] | 2024 | 0.9730 | 0.8049 | 0.9867 | 0.7894 | – |
| | Proposed (Fov) * | 2024 | 0.9687 | **0.9031** | 0.9802 | 0.8358 | 0.9891 |
| | Proposed (All) | 2024 | **0.9816** | 0.9029 | <u>0.9898</u> | 0.8358 | **0.9924** |

*Field of view (Fov) is defined for performance measurement. Fov denotes that only the internal pixels of the fundus are used for performance evaluation, while All indicates that all pixels in the images are taken into consideration.

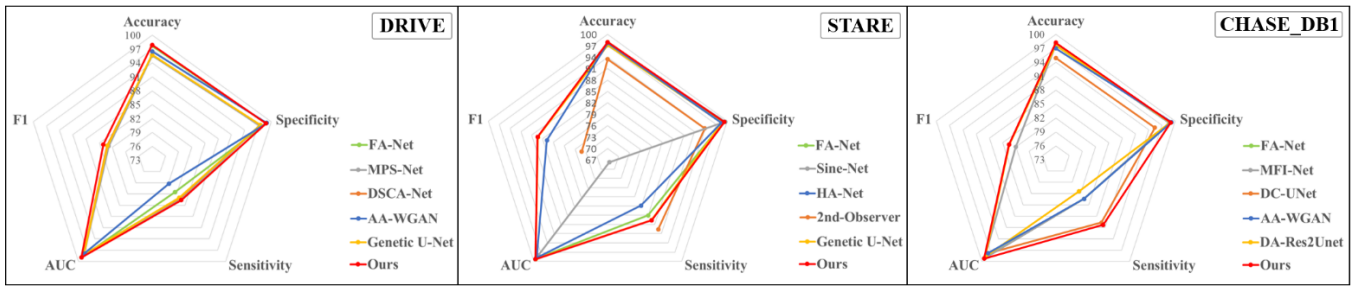

**Fig. 8.** The radar chart shows the performance of algorithms achieving the best or second-best results in key metrics across the three datasets. This comparison highlights the top-performing algorithms across these five critical indicators. The red line denotes the performance of the proposed MDFI-Net.

Among the introduced evaluation metrics, AUC is a significant performance indicator, especially when applied to image segmentation tasks. Unlike other metrics, AUC can assess the segmentation quality of probability maps by using different thresholds, and provides a more comprehensive evaluation result compared to single-threshold metrics. Besides, the F1 score is often used to analyze segmentation performance, particularly in cases of class imbalance, which is especially relevant in retinal vessel segmentation tasks, where the number of background pixels in fundus images greatly exceed the vessel pixels. For all datasets, the proposed method demonstrates superior performance in two comparative metrics (AUC and F1), which indicates that the constructed network is optimal in overall performance. Additionally, the proposed method also outperforms other methods in SEN (sensitivity), which suggests that the network is more effective in detecting vessels. In fundus images, non-vessel regions are significantly larger than vessel regions. Therefore, ACC (accuracy) is more persuasive than SPE (specificity), as it strikes a balance between SEN (sensitivity) and SPE (specificity). The ACC of proposed method is also superior than other methods. From TABLE V, it can be observed that the proposed method performs slightly lower on the SPE index compared to the AA-WGAN method. We believe that this may be due to the AA-WGAN method being more sensitive to TN, thus increasing its performance on the SPE index. It also can be observed that, the manual segmentation result of the second observer demonstrates significant performance on the SEN index. This is due to the different segmentation standards of blood vessels, which results in over segmentation on the fundus image. That is, a large number of blood vessels have been accurately segmented, but correspondingly, many background pixels have been mistakenly classified as blood vessels.

Since the state-of-the-art achieves outstanding performance in only one evaluation indicator, we conducted a comparative analysis between our method and the second-best approach on the radar chart, as depicted in Fig. 8. The results, from left to right, represent performance on the DRIVE, STARE, and CHASE_DB1 datasets. From the graph, it can be seen that the proposed method demonstrates outstanding performance across all metrics, rather than excelling in only one while lagging behind in others. Moreover, the results we obtained exhibit consistent performance across all three datasets without significant fluctuations. These figures illustrate the proposed MDFI-Net solves the inherent problems of the main methods in the field of retinal vessel segmentation and achieves performance improvement.

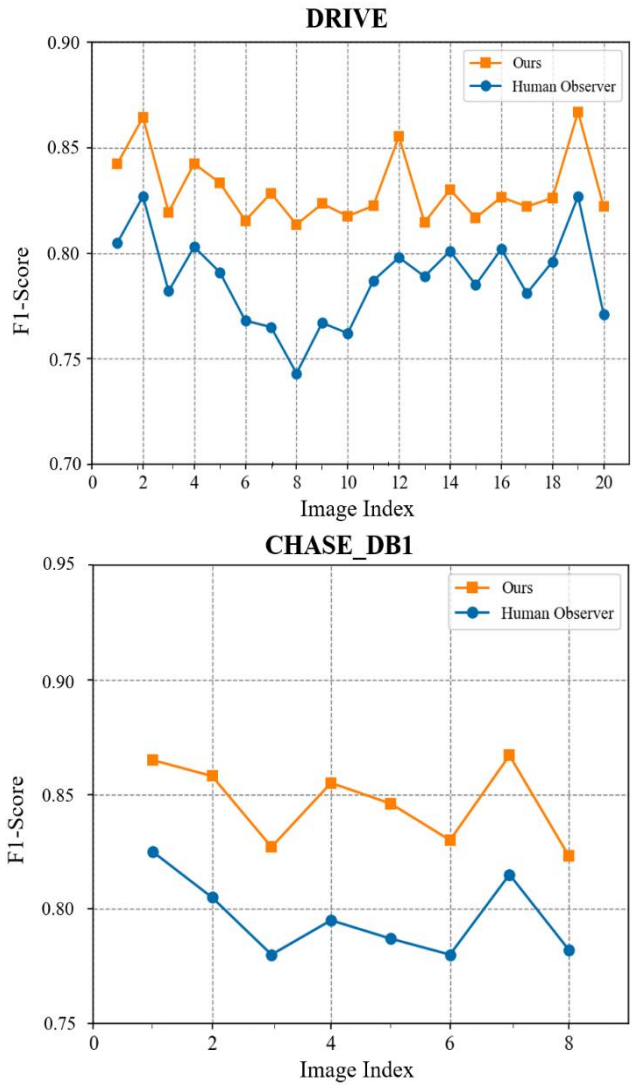

**Fig. 9** F1-Score of the proposed method and manual annotation on each image in the DRIVE and CHASE_DB1 test set.

In the retinal fundus image datasets, the test sets are annotated by two different experts. We use the annotations from the first expert as the ground truth, while the second expert's annotations are used as the human observer's result. One way to evaluate the quality of the segmentation results is to compare them with the results of a human observer. Fig. 9 shows the F1-Score of the proposed method and human observers on the DRIVE and CHASE_DB1 test sets. Compared with manual annotations, the segmentation results of the proposed method are closer to the ground truth for each test image. This indicates that our method can provide valuable segmentation results in practice. Although the expertise of these two experts provided important references for our research, there are still differences in expertise and annotator bias. In future research, we plan to collaborate with relevant hospitals to obtain annotations from more experts and further investigate the differences between the proposed method and expert annotations.

In order to visually demonstrate the segmentation results of different methods, we compare it with state-of-the-art methods on three public datasets (DRIVE, STARE, CHASE_DB1), including U-Net, CE-Net and IterNet. Fig. 10 shows the visual effects of our method and the above state-of-the-art methods on typical images of the three datasets. In the rectangular regions marked with a red line, we have provided an enlarged view of the densely distributed micro-vessels and the relatively robust main vessels, it can be observed that the proposed model is likely to provide more continuous segmentation boundaries and successfully segment micro-vessels. It indicates that our proposed network effectively reduces the instances of false-segmentation and missed-segmentation, while the irregular and multi-scale vessel structures also be well preserved. These visualizations further demonstrate the importance of aggregating multi-scale contextual information and capturing long-range dependencies in retinal vessel segmentation.

## VI. DISCUSSION

### A. Performance of Proposed DPCN

Due to the low contrast of tiny blood vessels in fundus images, directly feeding them into segmentation networks often leads to the target information being obscured by background. Therefore, we designed a Deformable-convolutional Pulse Coupling Network (DPCN) to address this issue. To demonstrate the feature extraction capability of proposed DPCN structure, we present a sequence of iterations in Fig. 11, where the resulting features are finally input into the segmentation network. As depicted in Fig. 11, (a) shows a color fundus image, with a magnified region containing the distribution of tiny blood vessels and its corresponding vessel labels; (b) presents the iterative results of the DPCN. As highlighted within the rectangular boxes, there is a branching vessel structure in the original image which is missed in the labels since the influence of low contrast. However, in the iterative results of the DPCN, this vessel structure is fully preserved. Additionally, the tiny structure within the circles is also retained in the iterative results of the DPCN, indicating the feature extraction capability of the proposed structure.

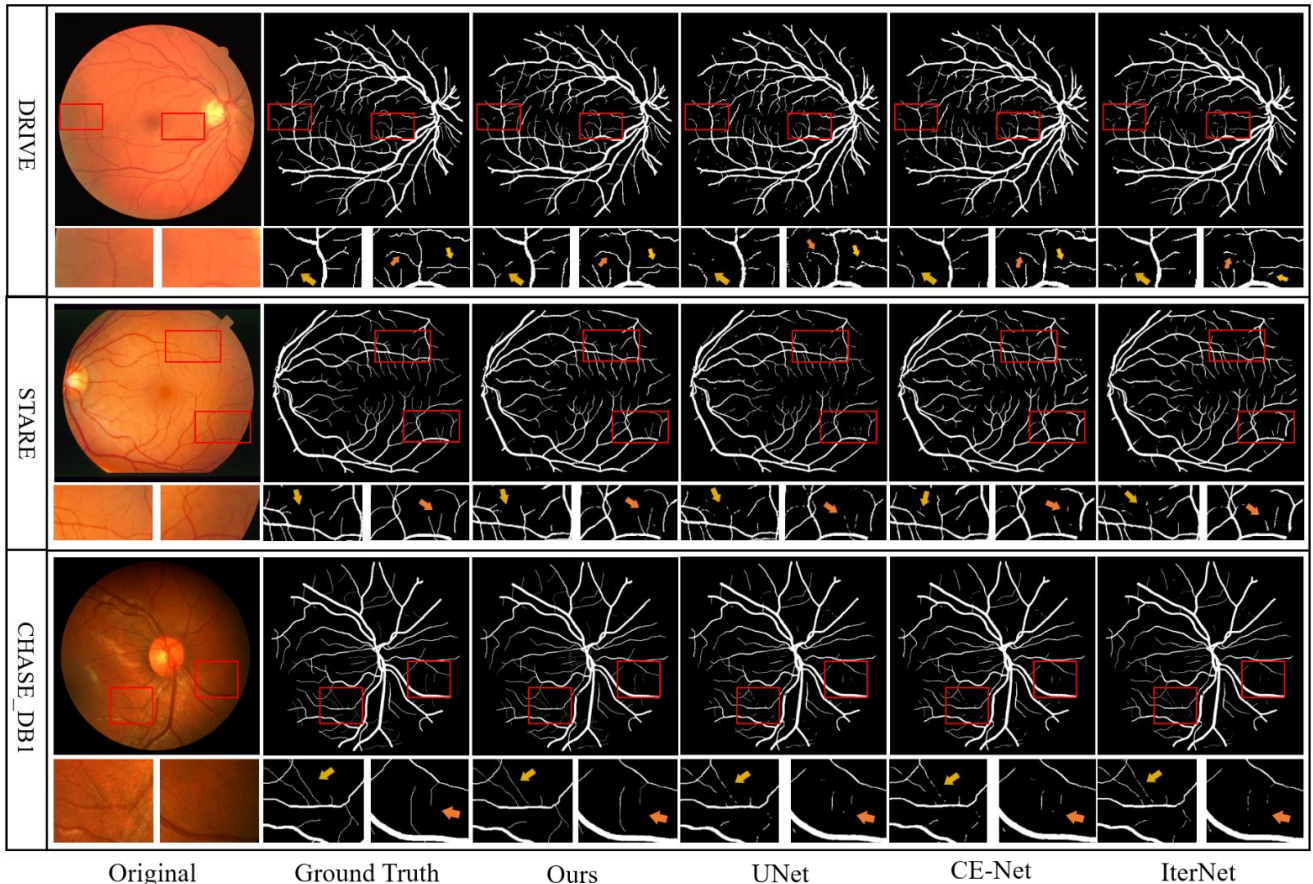

**Fig. 10.** Overall view visualization of the segmentation results.

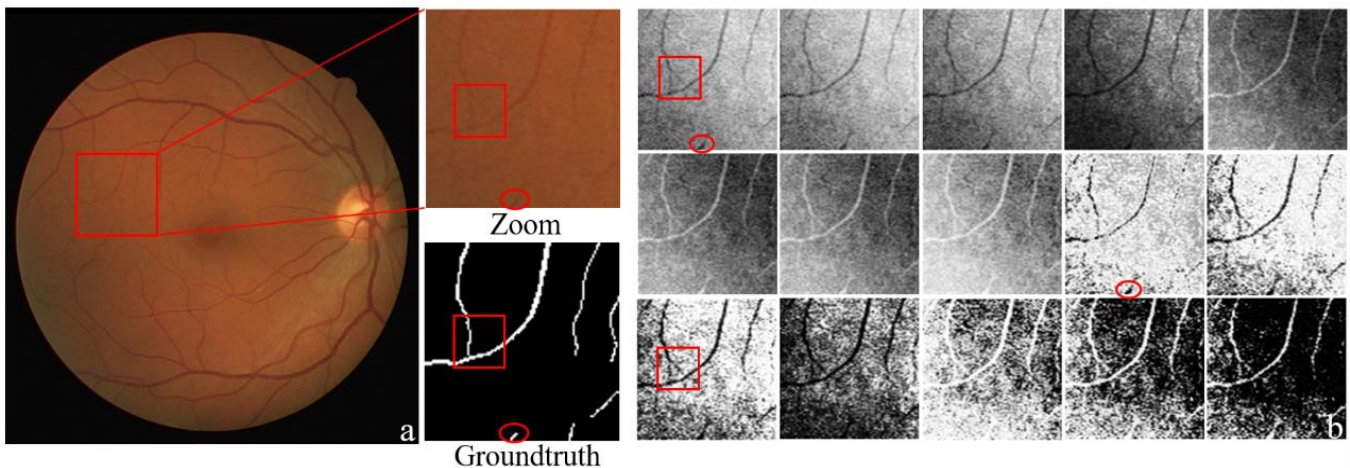

**Fig. 11.** The enhancement effect of DPCN structure on tiny retinal vessels. (a) is an original fundus image, with a local enlargement of tiny retinal vessels and corresponding label. (b) shows the iterative results of DPCN, in which the structural features of the retinal vessels are enhanced.

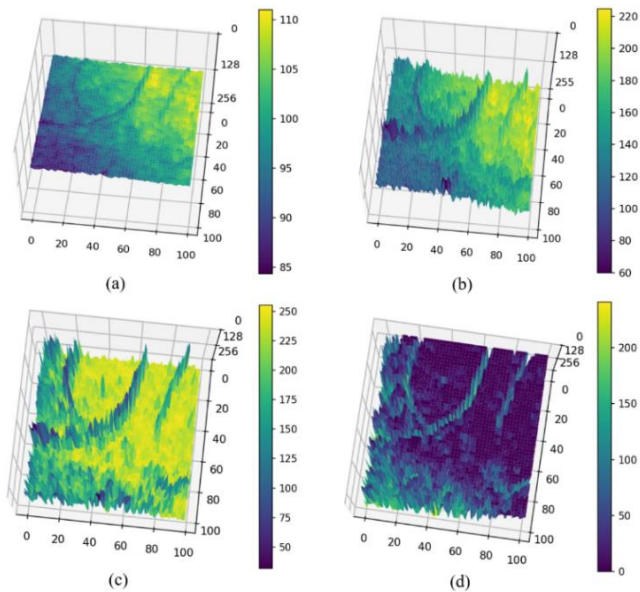

**Fig. 12.** The 3D space projection of grayscale images. (a) represents the original image displayed in 3D space, while (b), (c), and (d) depict the 1st, 10th, and 15th results of the DPCN iteration.

TABLE VIII

THE ABSOLUTE VALUES OF THE AVERAGE GRAYSCALE DIFFERENCES BETWEEN VASCULAR AND BACKGROUND REGIONS

| Iterations | original | 1st | 10th | 15th |
|---|---|---|---|---|
| Absolute Difference | 7.36 | 48.57 | 116.83 | 103.57 |

To visually demonstrate the enhancing effect of DPCN on vascular structures, we selected partial results from Fig. 11 and projected them into three-dimensional space for comparison, as illustrated in Fig 12. Here, Fig 12 (a) represents the original image displayed in three-dimensional space, while Fig 12 (b), Fig 12 (c), and Fig 12 (d) depict the 1st, 10th, and 15th iterations of the results from Fig. 11 (b), respectively. Additionally, we displayed the absolute values of the average grayscale differences between vascular and background regions in TABLE VIII. These findings demonstrate that the discovered architecture significantly enhances the grayscale contrast between target vessels and background, thus preserving vascular information in deeper layers of the network.

### B. Segmentation Performance on Different Scale Vessels

In Figure 13, we compared the results with the basic model on vascular structures of different thicknesses. In order to visually verify the effectiveness of the proposed model in vascular segmentation at different scales, we mapped the segmentation results onto grayscale images and labeled the segmentation results with different colors. In this representation, blue pixels indicate false negatives (FN), which correspond to vessel regions that were not detected; red pixels indicate false positives (FP), which are background regions incorrectly predicted as vessels; green pixels indicate true positives (TP), corresponding to correctly detected vessels. It is evident that the results of UNet and UNet++ exhibit more blue and red pixels, indicating a higher risk of under-segmentation and over-segmentation. This further shows the limitations of UNet and UNet++ in extracting and preserving features of small targets, while the discovered architecture excels over them. We observed that occlusions in the original images (such as the optic disc) increase the risks of both over-segmentation and under-segmentation in the results. The reduction of blue pixels in the segmentation results intuitively indicates significant improvements in addressing the under-segmentation issue with our proposed method, which we attribute to the contribution of the DPCN structure on the enhancement of blood vessel features. Furthermore, compared to Unet, the skip connections designed in Unet++ reduce the risk of over-segmentation and enhance the accuracy of segmentation of tiny vessel structures. This phenomenon is also reflected in our designed nested skip connection structure and is further improved by the introduced MSU unit.

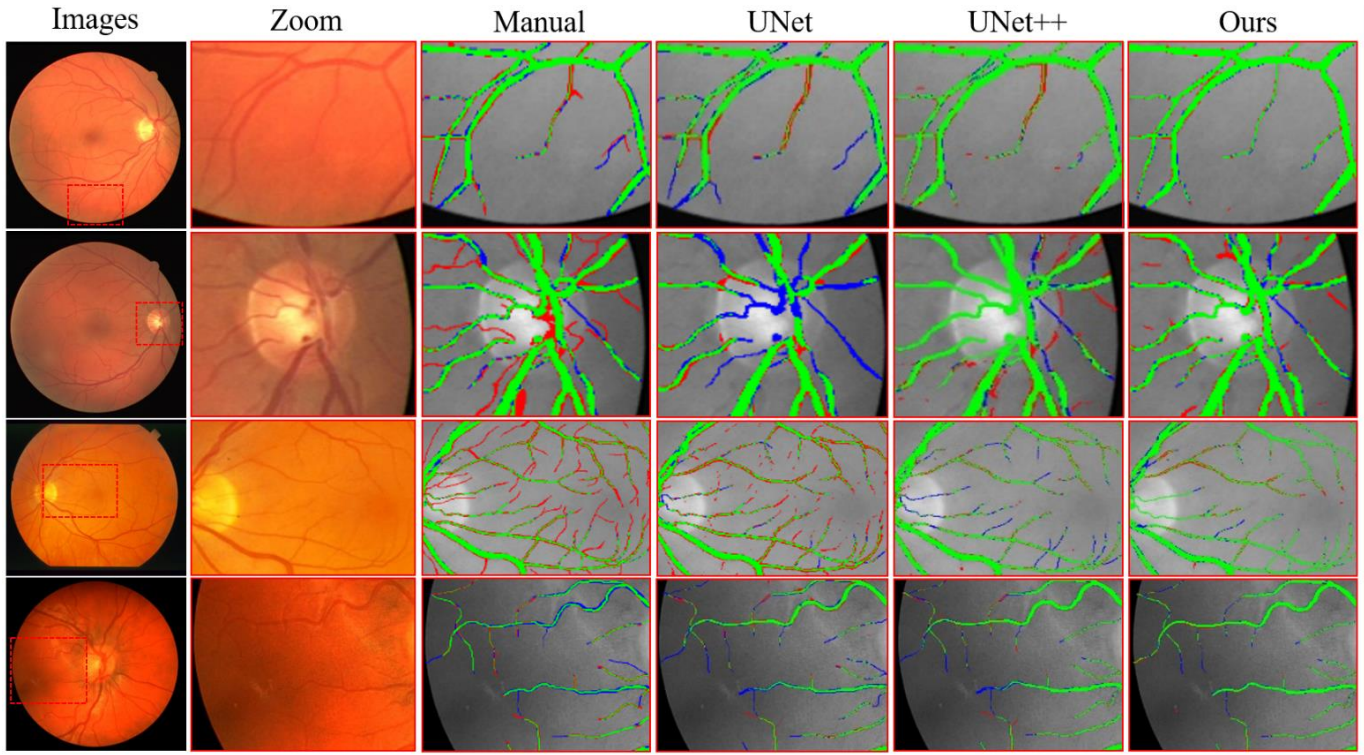

**Fig. 13.** Locally magnified view visualization of the segmentation results.

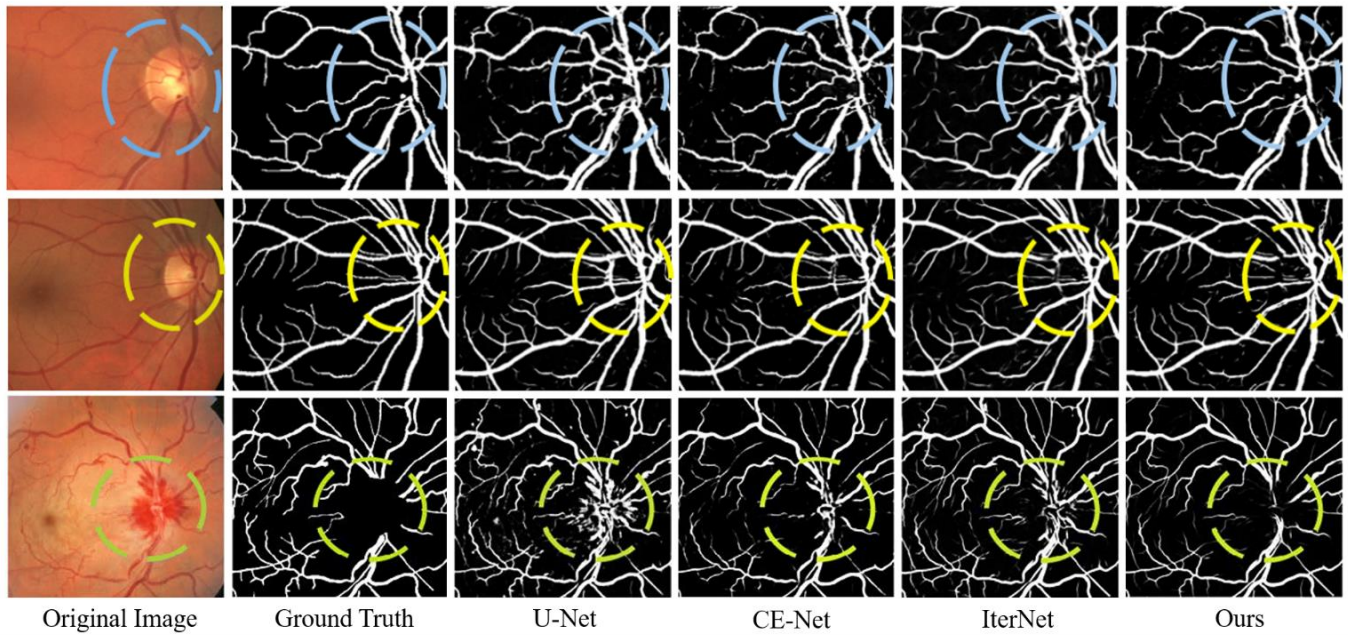

**Fig. 14.** Retinal vessel segmentation visualization on

### C. Interference from Obstructions

In fundus images, retinal vessel segmentation is often affected by obstructions, such as the structure and lesions of retinal organs. To further evaluate the capability of our proposed method, we selected images with occlusion interference in the DRIVE and STARE datasets for segmentation, the examples are shown in Fig. 14。Note that, with the compared methods, it is possible to inaccurately identify obstructions as vessels, leading to a certain degree of over segmentation. In comparison, our method can minimize the impact of obstructions, reduce the risk of over segmentation, and segment blood vessels more accurately.

### D. Limitation and Future Work

It is worth noting that although MDFI-Net demonstrates considerable effectiveness on retinal vessel segmentation across three public datasets, it still has some limitations. One major issue is that different datasets may require distinct preprocessing approaches, the preprocessing strategies we utilized might introduce variations in data distribution, potentially impacting the model's generalization ability. Consequently, further evaluation is required to determine the model's performance on more complex datasets. Additionally, in the realm of retinal vessel segmentation, supervised learning dominates, necessitating a substantial amount of high-quality data, which is often limited in clinical settings. Integrating our proposed model with unsupervised or weakly supervised methods may help alleviate this challenge. In future work, we plan to investigate the extraction of retinal vessel topology in 3D datasets, classification of retinal arteries and veins, and retinal image registration based on MDFI-Net, aiming to enhance disease diagnosis support.

## VII. CONCLUSION

In light of the inherent challenges in retinal vessel segmentation, a novel network named Multiscale Differential Feature Interaction Network (MDFI-Net) is proposed for this task. Current CNN models for retinal vessel segmentation face difficulties in significantly improving final segmentation performance, primarily due to the loss and overshadowing of low-contrast features in the deeper layers of the network. MDFI-Net introduces a Deformable-convolutional Pulse Coupling Network (DPCN) structure, which serves as an enhancement scheme for target vessel features. This structure enhances the feature of low-contrast micro-vessels and provides a series of complementary feature maps to the segmentation network iteratively. Additionally, the redesigned nested skip connections embedded with Multi-scale Subtraction Unit (MSU) facilitate cross-layer information exchange, further mitigating semantic loss during the decoding process. Extensive experiments validate the effectiveness of the proposed MDFI-Net, demonstrating its superior segmentation performance compared to existing state-of-the-art deep learning methods. Consequently, MDFI-Net offers a promising and effective solution for retinal vessel segmentation, aiding medical professionals in diagnosing related diseases through fundus images. We hope that this work can contribute to the assessment of vascular morphology and the diagnosis of ocular diseases.

## REFERENCES

[1] M. Seewoodhary, "An overview of diabetic retinopathy and other ocular complications of diabetes mellitus," Nurs Stand., vol. 36, no. 7, pp. 71-76, 2021. https://doi.org/10.7748/ns.2021.e11696

[2] J. Mo and L. Zhang, "Multi-level deep supervised networks for retinal vessel segmentation," Int. J. Comput. Assist. Radiol. Surg., vol. 12, no. 12, pp. 2181-2193, 2017. https://doi.org/10.1007/s11548-017-1619-0

[3] L. Benson, S. Yan, and H. Yan, "A novel vessel segmentation algorithm for pathological retina images based on the divergence of vector fields," IEEE Trans. Med. Imag., vol. 27, no. 2, pp. 237-246, Feb. 2008. https://doi.org/10.1088/1361-6560/acb98a

[4] J. Zhang, B. Dashtbozorg, E. Bekkers, J.P. Pluim, R. Duits, and B.M. ter Haar Romeny, "Robust retinal vessel segmentation via locally adaptive derivative frames in orientation scores," IEEE Trans. Med. Imag., vol. 35, no. 12, pp. 2631-2644, Dec. 2016. https://doi.org/10.1109/TMI.2016.2587062

[5] X. Zhang et al., "T-Net: Hierarchical Pyramid Network for Microaneurysm Detection in Retinal Fundus Image", IEEE Transactions on Instrumentation and Measurement, vol. 72, pp. 1-13, 2023, Art no. 5019613, https://doi.org/10.1109/TIM.2023.3286003

[6] A.M. Mendonca, and A. Campilho, "Segmentation of retinal blood vessels by combining the detection of centerlines and morphological reconstruction," IEEE Trans. Med. Imag., VOL. 25, NO. 9, pp. 1200-1213, 2016. https://doi.org/10.1109/TMI.2006.879955

[7] M. Monemian and H. Rabbani, "A Computationally Efficient Red-Lesion Extraction Method for Retinal Fundus Images", IEEE Transactions on Instrumentation and Measurement, vol. 72, pp. 1-13, 2023, Art no. 5001613. https://doi.org/10.1109/TIM.2022.3229712

[8] M. Tajmirriahi, R. Kafieh, Z. Amini and V. Lakshminarayanan, "A Dual-Discriminator Fourier Acquisitive GAN for Generating Retinal Optical Coherence Tomography Images", IEEE Transactions on Instrumentation and Measurement, vol. 71, pp. 1-8, 2022, Art no. 5015708, https://doi.org/10.1109/TIM.2022.3189735

[9] Z. Yu, and K. Sun. "Vessel segmentation on angiogram using morphology driven deformable model", in BMEI, Vol. 2, pp. 675-678, 2010. https://doi.org/10.1109/BMEI2010.5640056

[10] Y. Zhang, W. Hsu, and M. Lee. "Detection of retinal blood vessels based on nonlinear projections," Journal of Signal Processing Systems, vol. 55, pp.103-112, 2009. https://doi.org/10.1007/s11265-008-0179-5

[11] S. Kozerke, R. Botnar, S. Oyre, M.B. Scheidegger, E.M. Pedersen and P. Boesiger, "Automatic vessel segmentation using active contours in cine phase contrast flow measurements," Journal of Magnetic Resonance Imaging, vol. 10, pp. 41-51, 1999.

[12] A. Lahiri, A.G. Roy, D. Sheet and P.K. Biswas, "Deep neural ensemble for retinal vessel segmentation in fundus images towards achieving label-free angiography," in EMBC, pp. 1340-1343, 2016. https://doi.org/10.1109/EMBC.2016.7590955

[13] B. Hassan, S. Qin, T. Hassan, R. Ahmed and N. Werghi, "Joint Segmentation and Quantification of Chorioretinal Biomarkers in Optical Coherence Tomography Scans: A Deep Learning Approach", IEEE Transactions on Instrumentation and Measurement, vol. 70, pp. 1-17, 2021, Art no. 2508817, https://doi.org/10.1109/TIM.2021.3077988

[14] D. Das, D. R. Nayak and R. B. Pachori, "CA-Net: A Novel Cascaded Attention-Based Network for Multistage Glaucoma Classification Using Fundus Images", IEEE Transactions on Instrumentation and Measurement, vol. 72, pp. 1-10, 2023, Art no. 2531110, https://doi.org/10.1109/TIM.2023.3322499

[15] J. Pan, J. Gong, M. Yu, J. Zhang, Y. Guo and G. Zhang, "A Multilevel Remote Relational Modeling Network for Accurate Segmentation of Fundus Blood Vessels", IEEE Transactions on Instrumentation and Measurement, vol. 71, pp. 1-14, 2022, Art no. 5020714, https://doi.org/10.1109/TIM.2022.3203114

[16] J. Zhang et al., "Self-Guided Adversarial Network for Domain Adaptive Retinal Layer Segmentation", IEEE Transactions on Instrumentation and Measurement, vol. 73, pp. 1-10, 2024, Art no. 3527110, https://doi.org/10.1109/TIM.2024.3440388

[17] H. Wu, W. Wang, J. Zhong, B. Lei, Z. Wen, and J. Qin, "SCS-net: A scale and context sensitive network for retinal vessel segmentation," Med. Image Anal., vol. 70, 2021, Art. no. 102025. https://doi.org/10.1016/j.media.2021.102025

[18] H. Wang et al., "OCTFormer: An Efficient Hierarchical Transformer Network Specialized for Retinal Optical Coherence Tomography Image Recognition", IEEE Transactions on Instrumentation and Measurement, vol. 72, pp. 1-17, 2023, Art no. 2532217, https://doi.org/10.1109/TIM.2023.3329106

[19] L.C. Chen, G. Papandreou, I. Kokkinos, K. Murphy and A.L. Yuille, "Deeplab: Semantic image segmentation with deep convolutional nets, atrous convolution, and fully connected crfs," IEEE transactions on pattern analysis and machine intelligence, vol. 40, no.4, pp. 834-848, 2017. https://doi.org/10.1109/TPAMI.2017.2699184

[20] H. Su, L. Gao, Z. Wang, Y. Yu, J. Hong and Y. Gao, "A Hierarchical Full-Resolution Fusion Network and Topology-Aware Connectivity Booster for Retinal Vessel Segmentation", IEEE Transactions on Instrumentation and Measurement, vol. 73, pp. 1-16, 2024, Art no. 2521616, https://doi.org/10.1109/TIM.2024.3411133

[21] K. Hu, S. Jiang, Y. Zhang, X. Li and X. Gao, "Joint-Seg: Treat Foveal Avascular Zone and Retinal Vessel Segmentation in OCTA Images as a Joint Task", IEEE Transactions on Instrumentation and Measurement, vol. 71, pp. 1-13, 2022, Art no. 4007113, https://doi.org/10.1109/TIM.2022.3193188

[22] F. Yu, V. Koltun, and T. Funkhouser, "Dilated Residual Networks," in

CVPR, pp. 636-644, 2017.

[23] L. Chen, Y. Zhu, G. Papandreou, F. Schroff and H. Adam, "Encoder-decoder with atrous separable convolution for semantic image segmentation," in ECCV, pp. 801-808, 2018.

[24] K. He, X. Zhang, S. Ren and J. Sun, "Deep Residual Learning for Image Recognition." in CVPR, pp. 770-778, 2016.

[25] J. Li, G. Gao, L. Yang, G. Bian and Y. Liu, "DPF-Net: A Dual-Path Progressive Fusion Network for Retinal Vessel Segmentation", IEEE Transactions on Instrumentation and Measurement, vol. 72, pp. 1-17, 2023, Art no. 2517817, https://doi.org/10.1109/TIM.2023.3277946

[26] K . Li, X. Qi, Y. Luo, Z. Yao, X. Zhou and M. Sun, "Accurate retinal vessel segmentation in color fundus images via fully attention-based networks," IEEE J. Biomed. Health Informat., vol. 25, no. 6, pp. 2071–2081, 2021. https://doi.org/10.1109/JBHI.2020.3028180

[27] D. Wang, A. Haytham, J. Pottenburgh, O. Saeedi and Y. Tao, "Hard attention net for automatic retinal vessel segmentation," IEEE J. Biomed. Health Informat., vol. 24, no. 12, pp. 3384–3396, 2020. https://doi.org/10.1109/JBHI.2020.3002985

[28] X. He et al., "Lightweight Retinal Layer Segmentation With Global Reasoning", IEEE Transactions on Instrumentation and Measurement, vol. 73, pp. 1-14, 2024, Art no. 2520214, https://doi.org/10.1109/TIM.2024.3400305

[29] Z. Zhou, M.M.R. Siddiquee, N. Tajbakhsh and J. Liang, "UNet++: Redesigning Skip Connections to Exploit Multiscale Features in Image Segmentation," IEEE Trans. Med. Imag., vol. 39, no. 6, pp. 1856-1867, 2020. https://doi.org/10.1109/TMI.2019.2959609

[30] H. Huang, L. Lin, R. Tong, H. Hu, Q. Zhang, Y. Iwamoto, X. Han, Y. Chen and J. Wu, "UNet 3+: A Full-Scale Connected UNet for Medical Image Segmentation," IEEE International Conference on Acoustics, Speech and Signal Processing (ICASSP), pp. 1055-1059, 2020. https://doi.org/10.1109/ICASSP40776.2020.9053405

[31] Y. Li et al., "Diffusion Probabilistic Learning With Gate-Fusion Transformer and Edge-Frequency Attention for Retinal Vessel Segmentation", IEEE Transactions on Instrumentation and Measurement, vol. 73, pp. 1-13, 2024, Art no. 2523513, https://doi.org/10.1109/TIM.2024.3420264

[32] A. Mosinska, P. Márquez-Neila, M. Koziński and P. Fua, "Beyond the pixel-wise loss for topology-aware delineation," in Proceedings of the IEEE Conference on Computer Vision and Pattern Recognition, pp. 3136–3145, 2018.

[33] S. Shit, J. C. Paetzold, A. Sekuboyina and I. Ezhov, "Cldice-a novel topology preserving loss function for tubular structure segmentation," in Proceedings of the IEEE/CVF Conference on Computer Vision and Pattern Recognition, pp. 16560–16569, 2021.

[34] X. Liu, J. Cao, S. Wang, Y. Zhang and M. Wang, "Confidence-Guided Topology-Preserving Layer Segmentation for Optical Coherence Tomography Images With Focus-Column Module", IEEE Transactions on Instrumentation and Measurement, vol. 70, pp. 1-12, 2021, Art no. 5005612. https://doi.org/10.1109/TIM.2020.3047430

[35] X. Hu, F. Li, D. Samaras and C. Chen, "Topology-preserving deep image segmentation," Advances in Neural Information Processing Systems, vol. 32, 2019.

[36] X. Hu, Y. Wang, F. Li, D. Samaras, C. Chen, "Topology-aware segmentation using discrete morse theory," in International Conference on Learning Representations, 2020.

[37] R. Damseh, P. Pouliot, L. Gagnon, S. Sakadzic, D. Boas, F. Cheriet and F. Lesage, "Automatic graph-based modeling of brain microvessels captured with two-photon microscopy," IEEE journal of biomedical and health informatics, vol. 23, no. 6, pp. 2551–2562, 2018. https://doi.org/10.1109/JBHI.2018.2884678

[38] Z. Gu, J. Cheng, H. Fu, K. Zhou, H. Hao, Y. Zhao, T. Zhang, S. Gao and J. Liu, "Ce-net: Context encoder network for 2d medical image segmentation," IEEE Transactions on Medical Imaging, vol. 38, no. 10, pp. 2281–2292, 2019. https://doi.org/10.1109/TMI.2019.2903562

[39] L. Mou, Y. Zhao, H. Fu, Y. Liu, J. Cheng, Y. Zheng, P. Su, J. Yang, L. Chen and A.F. Frangi, "Cs²-net: Deep learning segmentation of curvilinear structures in medical imaging," Medical Image Analysis, vol. 67, p. 101874, 2021. https://doi.org/10.1016/j.media.2020.101874

[40] M. Tajmirriahi, R. Kafieh, Z. Amini and H. Rabbani, "A Lightweight Mimic Convolutional Auto-Encoder for Denoising Retinal Optical Coherence Tomography Images", IEEE Transactions on Instrumentation and Measurement, vol. 70, pp. 1-8, 2021, Art no. 4503908, https://doi.org/10.1109/TIM.2021.3072109

[41] F. Ding, G. Yang, J. Wu, D. Ding, J. Xv, G. Cheng and X. Li, "High-order attention networks for medical image segmentation," in International Conference on Medical Image Computing and Computer Assisted Intervention. Springer, pp. 253–262, 2020. https://doi.org/10.1007/978-3-030-59710-8_25

[42] Y. Ma, H. Hao, J. Xie, H. Fu, J. Zhang, J. Yang, Z. Wang, J. Liu, Y. Zheng, Y. Zhao, "Rose: A retinal oct-angiography vessel segmentation dataset and new model," IEEE Transactions on Medical Imaging, vol. 40, no. 3, pp. 928–939, 2020. https://doi.org/10.1109/TMI.2020.3042802

[43] Y. Li, Y. Zhang, W. Cui, B. Lei, X. Kuang and T. Zhang, "Dual encoder-based dynamic-channel graph convolutional network with edge enhancement for retinal vessel segmentation," IEEE Transactions on Medical Imaging, 2022. https://doi.org/10.1109/TMI.2022.3151666

[44] M. Zhang, F. Yu, J. Zhao, L. Zhang and Q. Li, "BEFD: Boundary enhancement and feature denoising for vessel segmentation," in International Conference on Medical Image Computing and Computer-Assisted Intervention. Springer, 2020, pp. 775–785.

[45] O. Ronneberger, P. Fischer, and T. Brox, "U-net: Convolutional networks for biomedical image segmentation," in Proc. Int. Conf. Med. Image Comput. Comput. Assist. Intervent. (MICCAI), Cham, Switzerland, 2015, pp. 234–241. https://doi.org/10.1007/978-3-319-24574-4_28

[46] E. O. Rodrigues, A. Conci, and P. Liatsis, E. O. Rodrigues, A. Conci, and P. Liatsis, "Element: Multi-modal retinal vessel segmentation based on a coupled region growing and machine learning approach," IEEE Journal of Biomedical and Health Informatics, vol. 24, no. 12, pp. 3507–3519, 2020. https://doi.org/10.1109/JBHI.2020.2999257

[47] Y. Xia, H. Hofmann, F. Dennerlein, K. Mueller, C. Schwemmer, S. Bauer, "Towards clinical application of a Laplace operator-based region of interest reconstruction algorithm in C-arm CT," IEEE Trans. Med. Imag., vol. 33, no. 3, pp. 593–606, Mar. 2014. https://doi.org/10.1109/TMI.2013.2291622

[48] H. Min, L. Hia, J. Han, X. Wang, Q. Pan, H. Fu, H. Wang, S. T. C. Wong and H. Li, "A multi-scale level set method based on local features for segmentation of images with intensity inhomogeneity," Pattern Recognition, 91 (2019): 69-85. https://doi.org/10.1016/j.patcog.2019.02.009

[49] Y. M. Kassim, O. V. Glinskii, V. V. Glinsky, V. H. Huxley, G. Guidoboni, K. Palaniappan, "Deep u-net regression and hand-crafted feature fusion for accurate blood vessel segmentation," in IEEE International Conference on Image Processing. pp. 1445–1449, 2019. https://doi.org/10.1109/ICIP.2019.8803084

[50] Y. Li, Z. Qiao, S. Zhang, Z. Wu, X. Mao, J. Kou and H. Qi, "A novel method for low-contrast and high-noise vessel segmentation and location in venipuncture," IEEE Transactions on Medical Imaging, vol. 36, no. 11, pp. 2216–2227, 2017. https://doi.org/10.1109/TMI.2017.2732481

[51] Z. Qu, S. Wang, L. Liu and D. Zhou, "Visual cross-image fusion using deep neural networks for image edge detection," IEEE Access, vol. 7, pp. 57604–57615, 2019. https://doi.org/10.1109/ACCESS.2019.2914151

[52] H. Huang, L. Lin, R. Tong, H. Hu, Q. Zhang, Y. Iwamoto, X. Han, Y. Chen and J. Wu, "WNET: An end-to-end atlas-guided and boundary enhanced network for medical image segmentation," in Proc. IEEE 17th Int. Symp. Biomed. Imag. (ISBI), pp. 763–766, 2020. https://doi.org/10.1109/ISBI45749.2020.9098654

[53] G. Huang, D. Chen, T. Li, F. Wu, L. Van Der Maaten and K.Q. Weinberger, "Multi-scale dense networks for resource efficient image classification," in Proc. Int. Conf. Learn. Representations, 2018. https://doi.org/10.48550/arXiv.1703.09844

[54] T.S. Sheikh, Y. Lee, and M. Cho, "Histopathological classification of breast cancer images using a multi-scale input and multi-feature network," Cancers, vol. 12, no. 2031, 2020. https://doi.org/10.3390/cancers12082031

[55] L. Qian, C. Wen, Y. Li, Z. Hu, X. Zhou, X. Xia and S. H. Kim, "Multi-scale context UNet-like network with redesigned skip connections for medical image segmentation," Computer Methods and Programs in Biomedicine, Vol. 243, 2024. https://doi.org/10.1016/j.cmpb.2023.107885

[56] C. Li, Y. Tan, W. Chen, X. Luo, Y. Gao, X. Jia and Z. Wang, "Attention unet++: A Nested Attention-Aware U-Net for Liver CT Image Segmentation," in 2020 IEEE international conference on image processing (ICIP), pp. 345–349, 2020. https://doi.org/10.1109/ICIP40778.2020.9190761

[57] R. Azad, M. Asadi-Aghbolaghi, M. Fathy, S. Escalera, "Bi-directional ConvLSTM U-Net with densely connected convolutions," In Proceedings of the IEEE/CVF international conference on computer vision (ICCV).

[58] H. Li, J. Fang, S. Liu, X. Liang, X. Yang, Z. Mai, M.T. Van, T. Wang, Z. Chen, D. Ni, "CR-Unet: A Composite Network for Ovary and Follicle Segmentation in Ultrasound Images," in IEEE Journal of Biomedical and Health Informatics, vol. 24, no. 4, pp. 974-983, 2020. https://doi.org/10.1109/JBHI.2019.2946092

[59] G.C. Ates, P. Mohan and E. Celik, Gorkem Can Ates, et al., "Dual Cross-Attention for medical image segmentation," Engineering Applications of Artificial Intelligence, Vol. 126, Part D, 2023. https://doi.org/10.1016/j.engappai.2023.107139

[60] G. Huang, Z. Liu, L. van der Maaten, K. Q. Weinberger, "Densely connected convolutional networks," in Proc. IEEE Conf. Comput. Vis. pattern Recognit., pp. 4700–4708, 2017.

[61] X. Deng, and J. Ye, X. Deng, and J. Ye, "A retinal blood vessel segmentation based on improved D-MNet and pulse-coupled neural network," Biomedical Signal Processing and Control, Vol. 73, 2022. https://doi.org/10.1016/j.bspc.2021.103467

[62] K. Upadhyay, M. Agrawal, P. Vashist, "Unsupervised multiscale retinal blood vessel segmentation using fundus images." IET Image Processing, vol. 14, no. 11, pp. 2616–2625, 2020. https://doi.org/10.1049/iet-ipr.2019.0969

[63] D.A. Palanivel, S. Natarajan and S. Gopalakrishnan, "Retinal vessel segmentation using multifractal characterization," Applied Soft Computing, vol. 94, 2020. https://doi.org/10.1016/j.asoc.2020.106439

[64] M. Z. Alom, C. Yakopcic, M. Hasan, T. M. Taha, and V. K. Asari, "Recurrent residual U-Net for medical image segmentation," J. Med. Imag., vol. 6, no. 1, Mar. 2019, Art. no. 014006. https://doi.org/10.1117/1.JMI.6.1.014006

[65] T.M. Khan, M.A.U. Khan, N.U. Rehman, K. Naveed, I.U. Afridi, S.S. Naqvi and I. Raazak, "Width-wise vessel bifurcation for improved retinal vessel segmentation." Biomedical Signal Processing and Control, vol. 71, 2022. https://doi.org/10.1016/j.bspc.2021.103169

[66] L. Li, M. Verma, Y. Nakashima, H. Nagahara and R. Kawasaki, "Iternet: Retinal image segmentation utilizing structural redundancy in vessel networks." Proceedings of the IEEE/CVF winter conference on applications of computer vision, 2020.

[67] C. Guo, M. Szemenyei, Y. Yi, W. Wang, B. Chen, and C. Fan, "SA-UNet: Spatial attention U-Net for retinal vessel segmentation," in Proc. 25th Int. Conf. Pattern Recognit. (ICPR), Jan. 2021, pp. 1236–1242. https://doi.org/10.1109/ICPR48806.2021.9413346

[68] Y. Cheng, M. Ma, L. Zhang, C. Jin, L. Ma and Y. Zhou, "Retinal blood vessel segmentation based on Densely Connected U-Net," Mathematical Biosciences and Engineering, vol. 17, pp. 3088–3108, 2020. https://doi.org/10.3934/mbe.2020175

[69] Y. Liu, J. Shen, L. Yang, H. Yu, and G. Bian, "Wave-net: A lightweight deep network for retinal vessel segmentation from fundus images," Comput. Biol. Med., vol. 152, Jan. 2023, Art. no. 106341. https://doi.org/10.1016/j.compbiomed.2022.106341

[70] Z. Lin, J. Huang, Y. Chen, X. Zhang, W. Zhao, Y. Li, L. Lu, M. Zhan, X. Jiang and X. Liang, "A high resolution representation network with multi-path scale for retinal vessel segmentation," Computer Methods and Programs in Biomedicine, vol. 208, 2021. https://doi.org/10.1016/j.cmpb.2021.106206

[71] J. Wei, G. Zhu, Z. Fan, J. Liu, Y. Rong, J. Mo, W. Li, X. Chen, J. Wei, G. Zhu, Z. Fan, et al., "Genetic U-Net: automatically designed deep networks for retinal vessel segmentation using a genetic algorithm," IEEE Transactions on Medical Imaging, vol. 41, no. 2, pp. 292–307, 2021. https://doi.org/10.1109/TMI.2021.3111679

[72] Y. Ye, C. Pan, Y. Wu, S. Wang, Y. Xia, "MFI-Net: Multiscale Feature Interaction Network for Retinal Vessel Segmentation," in IEEE Journal of Biomedical and Health Informatics, vol. 26, no. 9, pp. 4551–4562, 2022. https://doi.org/10.1109/JBHI.2022.3182471

[73] S. Guo, "CSGNet: Cascade semantic guided net for retinal vessel segmentation," Biomedical Signal Processing and Control, vol. 78, 2022. https://doi.org/10.1016/j.bspc.2022.103930

[74] T.M. Khan, S.S. Naqvi, A. Robles-Kelly, I. Razzak, "Retinal vessel segmentation via a Multi-resolution Contextual Network and adversarial learning," Neural Networks, vol. 165, pp. 310–320, 2023. https://doi.org/10.1016/j.neunet.2023.05.029

[75] R. Liu, T. Wang, X. Zhang and X. Zhou, "DA-Res2UNet: Explainable blood vessel segmentation from fundus images," Alexandria Engineering Journal, vol. 68, pp. 539–549, 2023. https://doi.org/10.1016/j.aej.2023.01.049

[76] M. Liu, Z. Wang, H. Li, P. Wu, F.E. Alsaadi and N. Zeng, "AA-WGAN: Attention augmented Wasserstein generative adversarial network with application to fundus retinal vessel segmentation." Computers in Biology and Medicine, vol. 158, 2023. https://doi.org/10.1016/j.compbiomed.2023.106874

[77] I. Atli and O.S. Gedik, Sine-Net: A fully convolutional deep learning architecture for retinal blood vessel segmentation, Engineering Science and Technology, an International Journal 24 (2) (2021) 271–283. https://doi.org/10.1016/j.jestch.2020.07.008

[78] Q. Jin, H. Hou, G. Zhang, H. Wang, and Z. Li, "Swin-ASNet: An adaptive RGB-selection network with Swin transformer for retinal vessel segmentation," in Proc. IEEE Int. Conf. Multimedia Expo (ICME), Brisbane, QC, Australia, Jul. 2023, pp. 1415–1420. https://doi.org/10.1109/ICME55011.2023.00245

[79] H. Zhang et al., "TiM-net: Transformer in M-net for retinal vessel segmentation," J. Healthcare Eng., vol. 2022, pp. 1–17, Jul. 2022, Art. no. 9016401 https://doi.org/10.1155/2022/9016401

[80] G. Gao, J. Li, L. Yang, and Y. Liu, "A multi-scale global attention network for blood vessel segmentation from fundus images," Measurement, vol. 222, Nov. 2023, Art. no. 113553. https://doi.org/10.1016/j.measurement.2023.113553

[81] H. Su, L. Gao, Z. Wang, Y. Yu, J. Hong and Y. Gao, "A Hierarchical Full-Resolution Fusion Network and Topology-Aware Connectivity Booster for Retinal Vessel Segmentation," in IEEE Transactions on Instrumentation and Measurement, vol. 73, pp. 1-16, 2024, Art no. 2521616. https://doi.org/10.1109/TIM.2024.3411133

[82] Mingtao Liu, Yunyu Wang, Lei Wang, et al, "IMFF-Net: An integrated multi-scale feature fusion network for accurate retinal vessel segmentation from fundus images", Biomedical Signal Processing and Control, Volume 91, 2024, 105980. https://doi.org/10.1016/j.bspc.2024.105980